\documentclass{article}

\usepackage[preprint,nonatbib]{neurips_data_2021}

\usepackage[utf8]{inputenc} 
\usepackage[T1]{fontenc}    
\usepackage{hyperref}       
\usepackage{url}            
\usepackage{booktabs}       
\usepackage{amsfonts}       
\usepackage{nicefrac}       
\usepackage{microtype}      
\usepackage[table,xcdraw]{xcolor}         
\usepackage{algorithmic}
\usepackage{graphicx}
\usepackage{multirow}
\usepackage{floatrow}
\usepackage{subcaption}
\usepackage{wrapfig}
\usepackage{paralist}

\title{\NAME: A Dataset Generator for a Systematic Evaluation of Adversarial Robustness  of Vision Models}

\author{%
  Federico~Nesti, Giulio Rossolini, Gianluca~D'Amico, Alessandro~Biondi, Giorgio~Buttazzo
    \\
  Department of Excellence in Robotics \& AI\\
  Scuola Superiore Sant'Anna, Pisa, Italy \\
  \texttt{name.surname@santannapisa.it} \\
}

\newcommand{\NAME}{\textsc{carla-GeAR}}

\newcommand{\colb}{0.32\textwidth}

\begin{document}

\maketitle

\begin{abstract}

Adversarial examples represent a serious threat for deep neural networks in several application domains and a huge amount of work has been produced to investigate them and mitigate their effects. Nevertheless, no much work has been devoted to the generation of datasets specifically designed to evaluate the adversarial robustness of neural models.
This paper presents \NAME, a tool for the automatic generation of photo-realistic synthetic datasets that can be used for a systematic evaluation of the adversarial robustness of neural models against physical adversarial patches, as well as for comparing the performance of different adversarial defense/detection methods. The tool is built on the CARLA simulator \cite{2017arXiv171103938D}, using its Python API, and allows the generation of datasets for several vision tasks in the context of autonomous driving. 
The adversarial patches included in the generated datasets are attached to billboards or the back of a truck and are crafted by using state-of-the-art white-box attack strategies to maximize the prediction error of the model under test. Finally, the paper presents an experimental study to evaluate the performance of some defense methods against such attacks, showing how the datasets generated with \NAME~might be used in future work as a benchmark for adversarial defense in the real world. All the code and datasets used in this paper are available at \url{http://carlagear.retis.santannapisa.it}. 
\end{abstract}

\section{Introduction} \label{s:intro}

\begin{figure}
\centering
\setlength\tabcolsep{0.1pt}
\renewcommand{\arraystretch}{0.8}
\begin{tabular}{ccccc}
&RGB image & Ground Truth label & Network Prediction& \\ \centering
\rotatebox[origin=l]{90}{\scriptsize ~~~~~~~Billboard04} & 
\begin{subfigure}[t]{\colb}
    \centering
    \includegraphics[width=\textwidth]{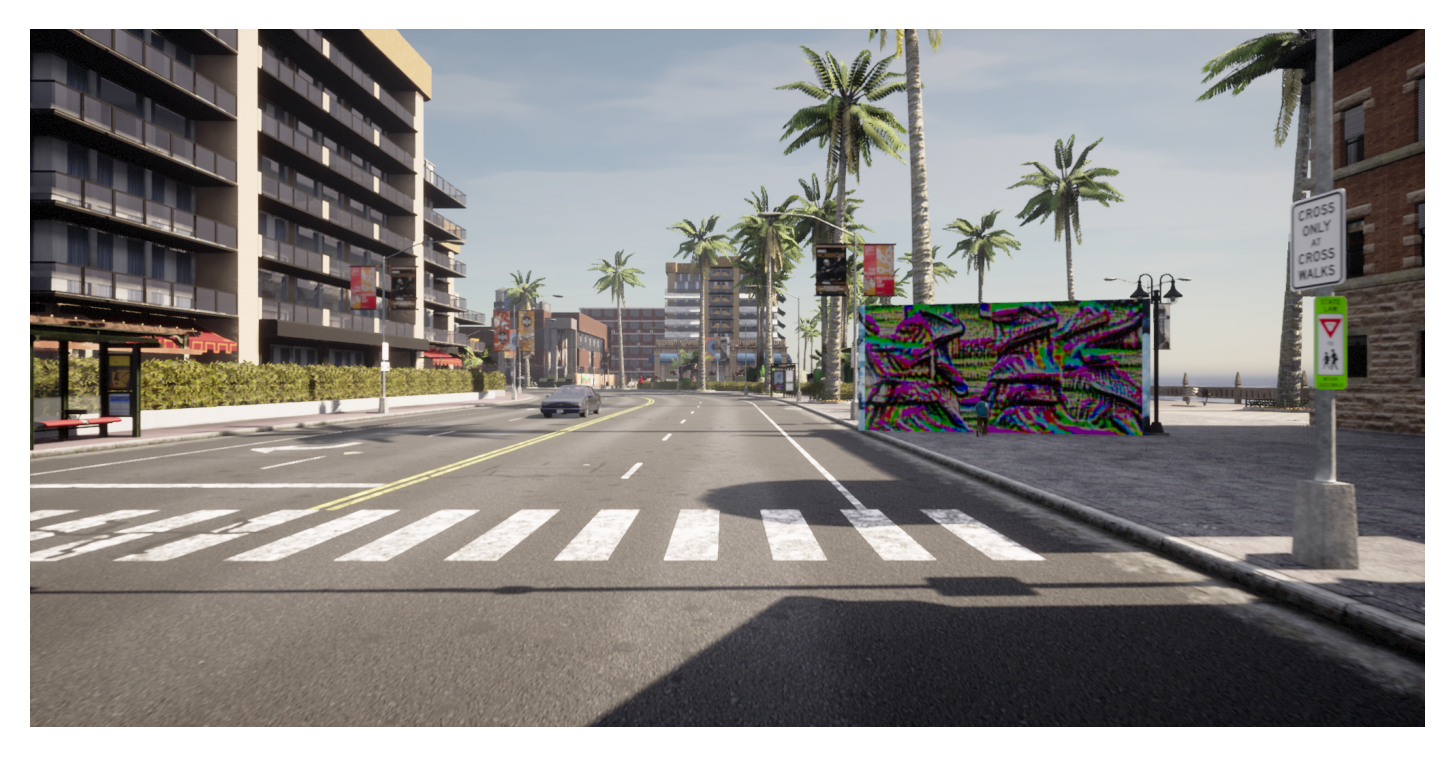}
\end{subfigure} &
\begin{subfigure}[t]{\colb}
    \centering
    \includegraphics[width=\textwidth]{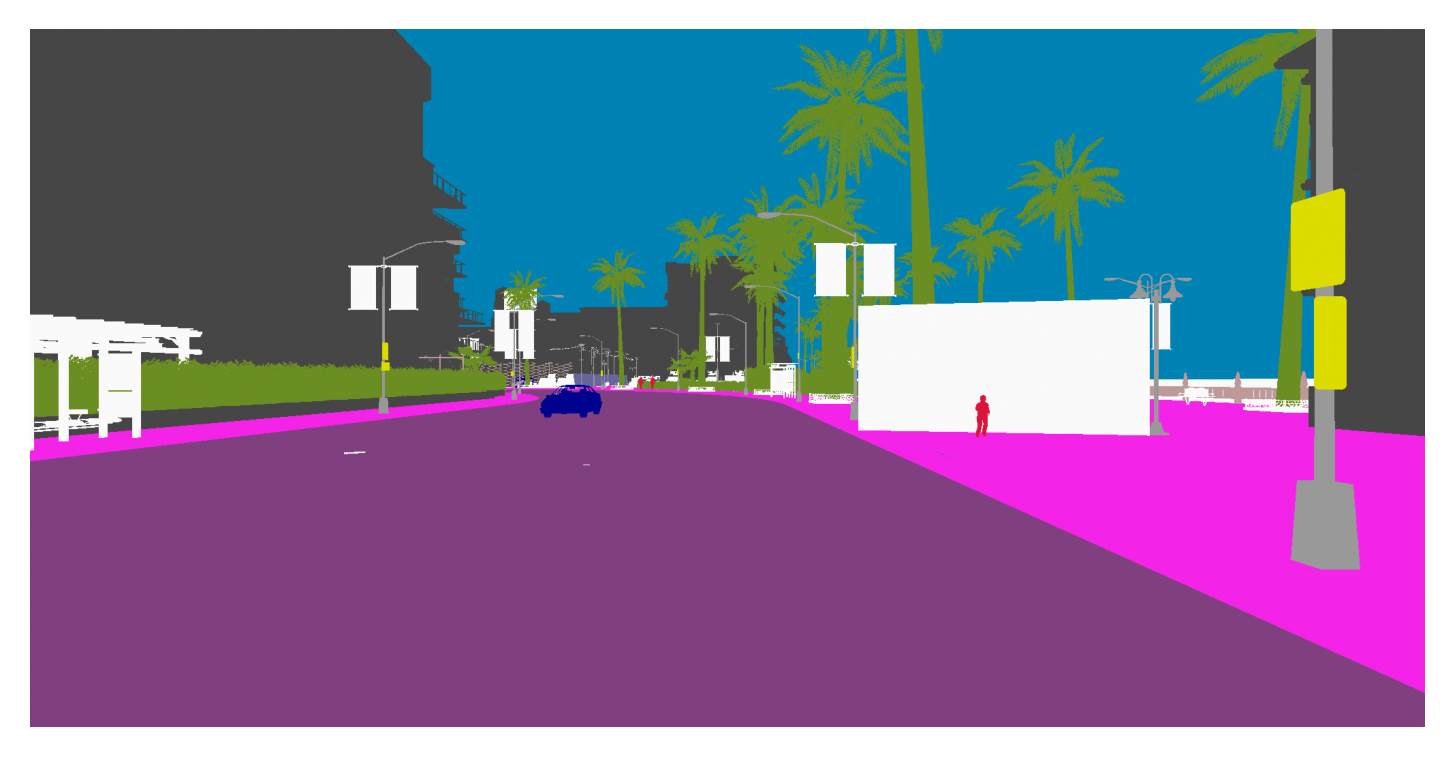}
\end{subfigure} &
\begin{subfigure}[t]{\colb}
    \centering
    \includegraphics[width=\textwidth]{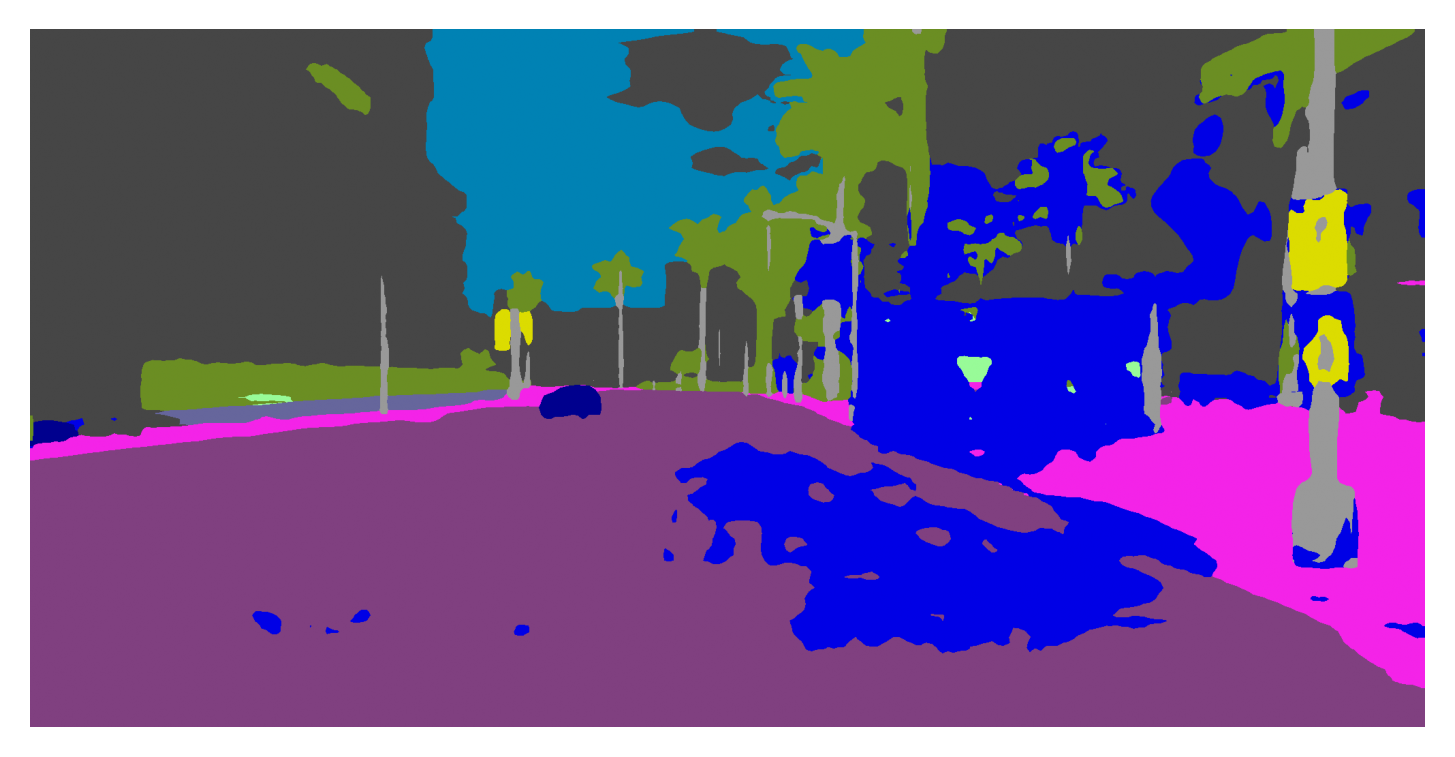}
\end{subfigure}& \rotatebox[origin=r]{-90}{\raggedleft \scriptsize  DDRNet~~~~~~~~}\\
\setlength\extrarowheight{-10pt}
\rotatebox[origin=l]{90}{\scriptsize ~~~~~~~~~Billboard05} & \begin{subfigure}[t]{\colb}
    \centering
    \includegraphics[width=\textwidth]{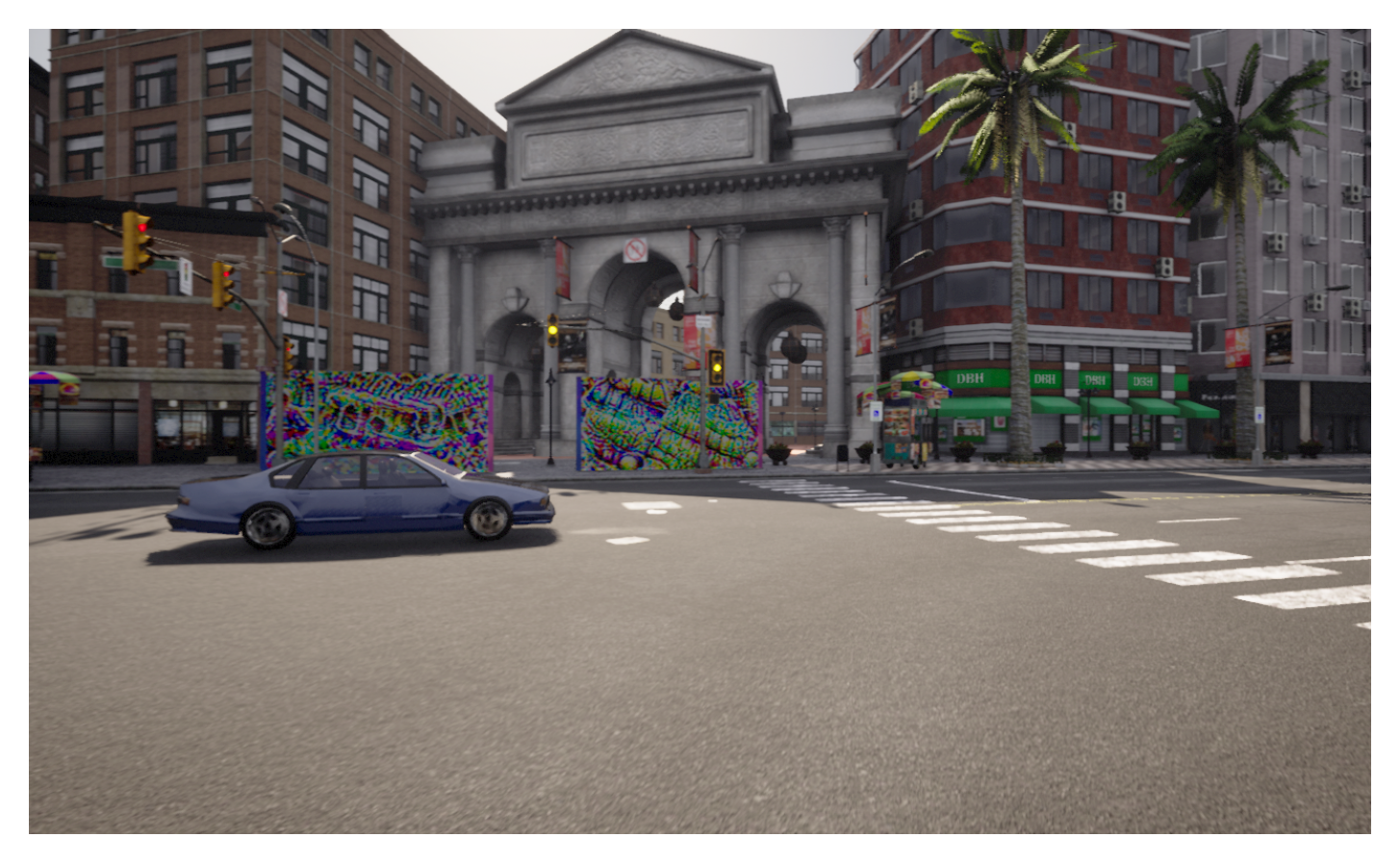}
\end{subfigure} &
\begin{subfigure}[t]{\colb}
    \centering
    \includegraphics[width=\textwidth]{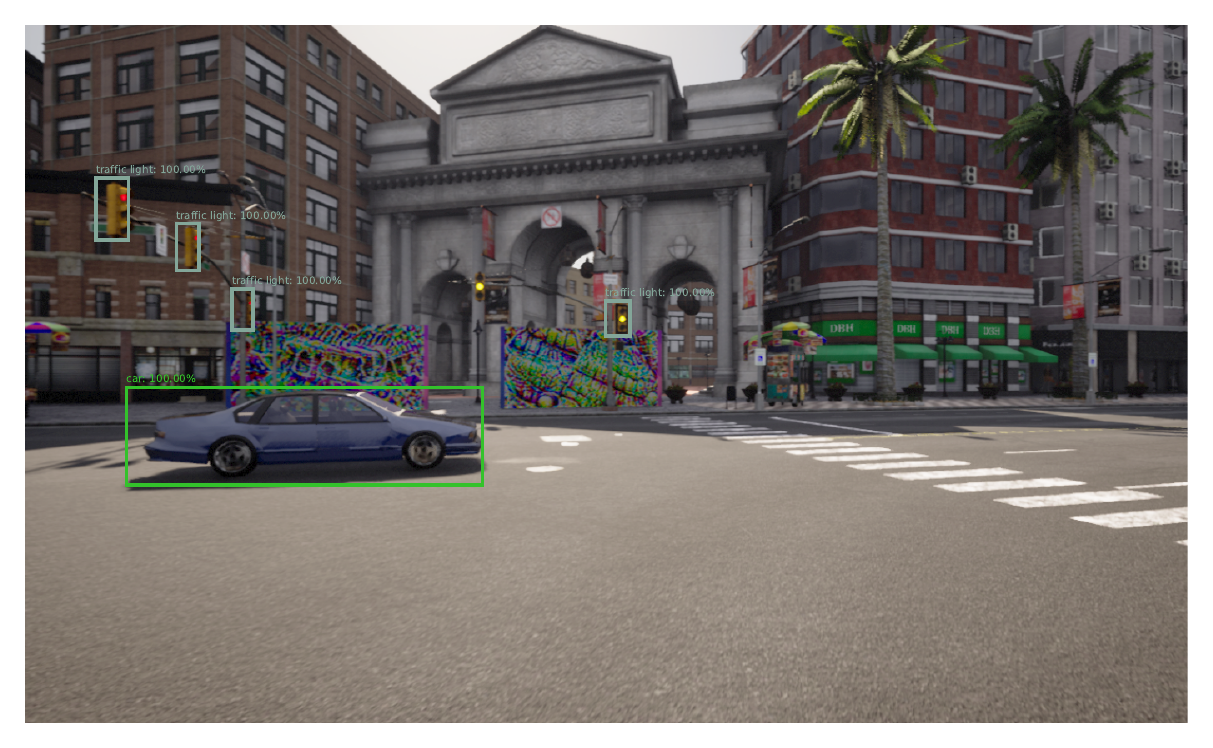}
\end{subfigure} &
\begin{subfigure}[t]{\colb}
    \centering
    \includegraphics[width=\textwidth]{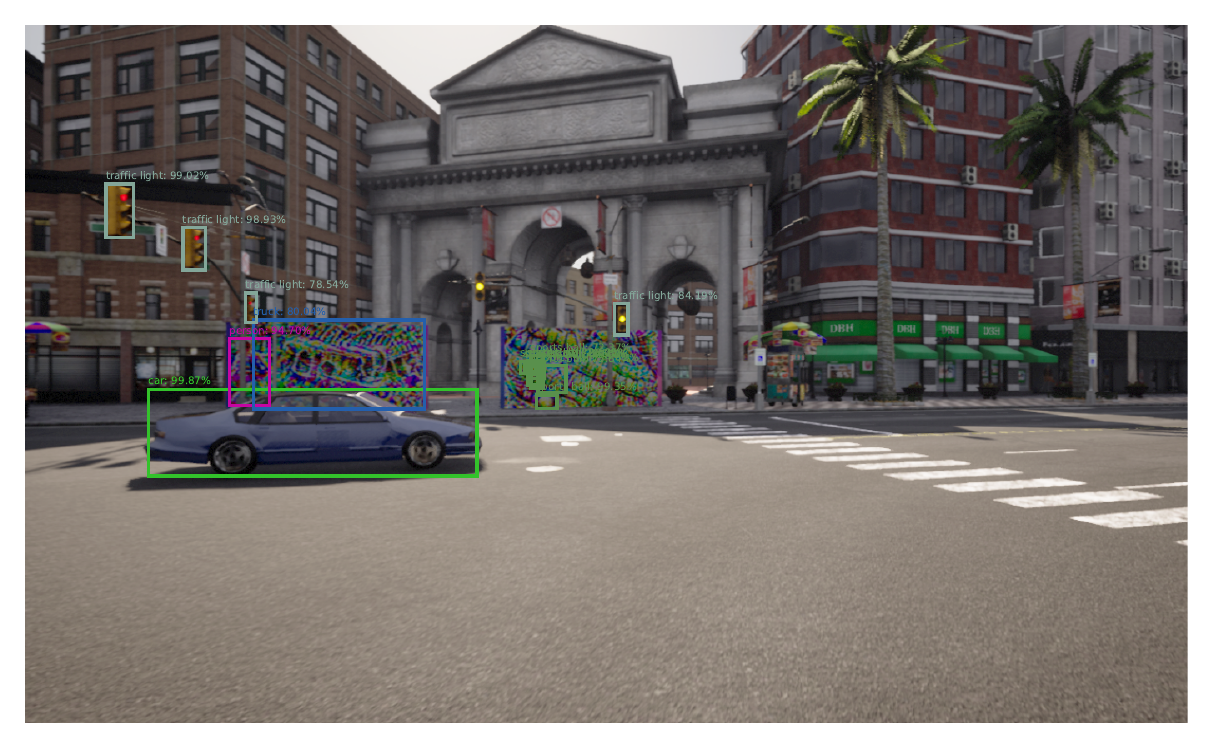}
\end{subfigure} &
\rotatebox[origin=r]{-90}{\raggedleft \scriptsize  Faster R-CNN~~~~~~~~~}\\
\rotatebox[origin=l]{90}{\scriptsize Billboard02}&\begin{subfigure}[t]{\colb}
    \centering
    \includegraphics[width=\textwidth]{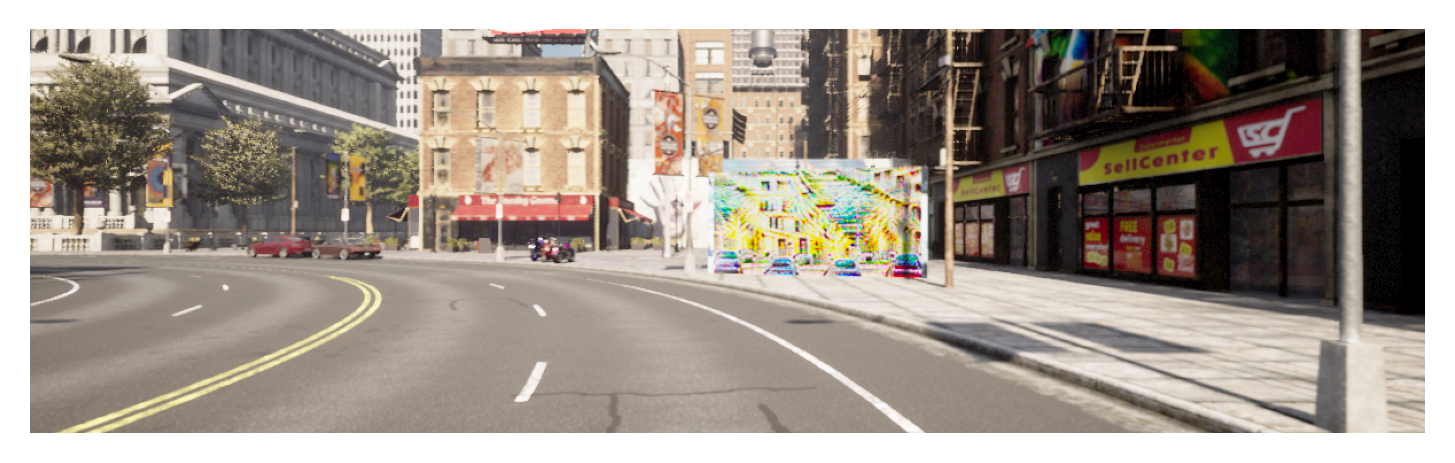}
\end{subfigure} &
\begin{subfigure}[t]{\colb}
    \centering
    \includegraphics[width=\textwidth]{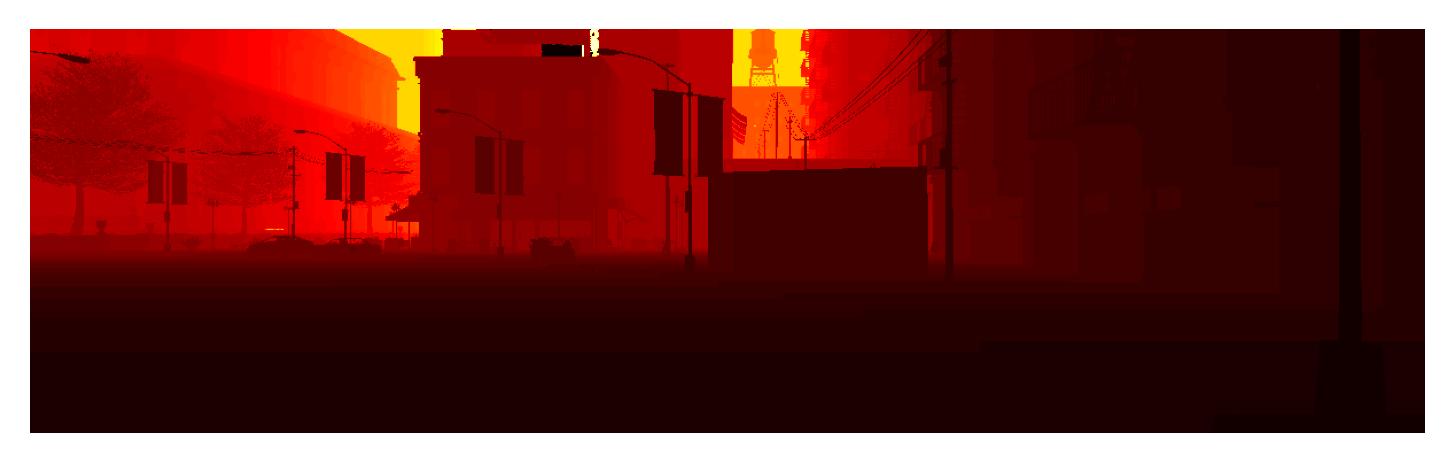}
\end{subfigure} &
\begin{subfigure}[t]{\colb}
    \centering
    \includegraphics[width=\textwidth]{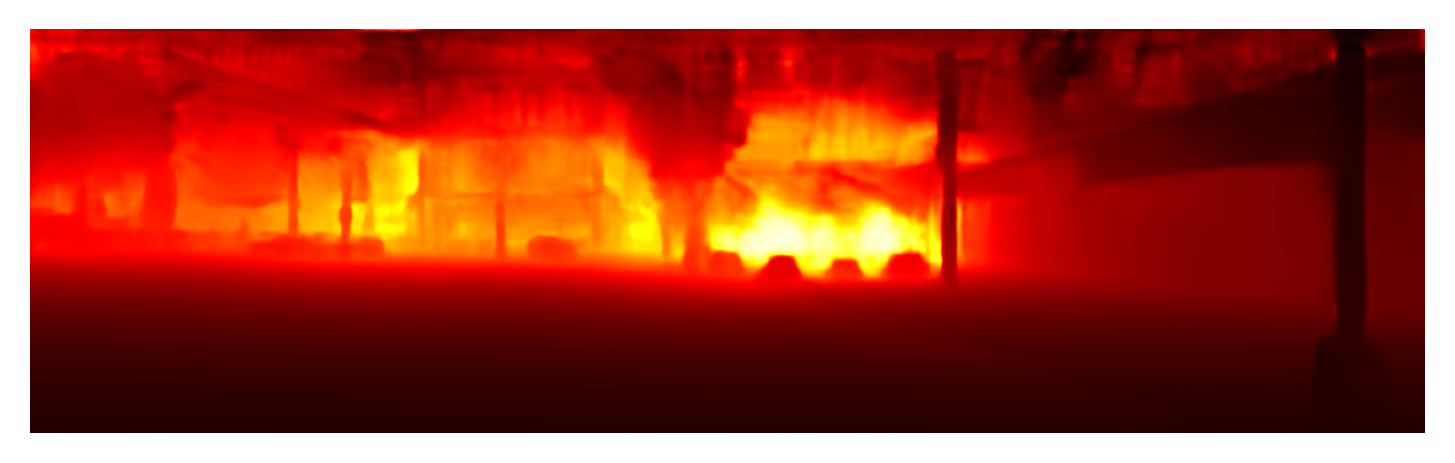}
\end{subfigure} &
\rotatebox[origin=r]{-90}{\raggedleft \scriptsize  GLPDepth~~}\\
\rotatebox[origin=l]{90}{\scriptsize ~~Billboard09}&\begin{subfigure}[t]{\colb}
    \centering
    \includegraphics[width=\textwidth]{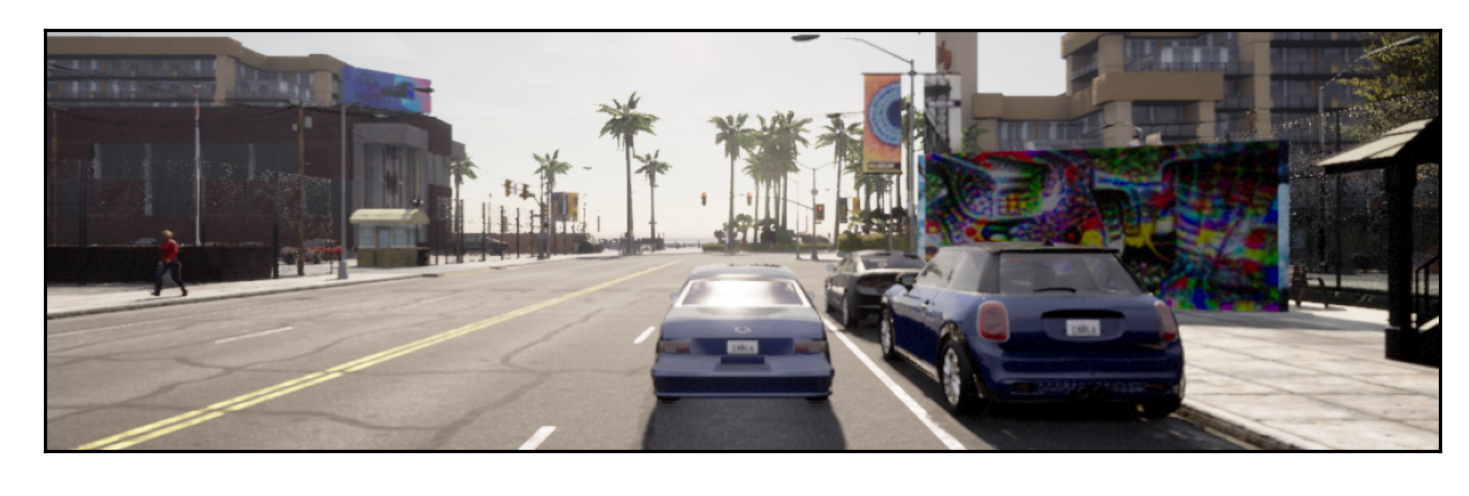}
\end{subfigure} &
\begin{subfigure}[t]{\colb}
    \centering
    \includegraphics[width=\textwidth]{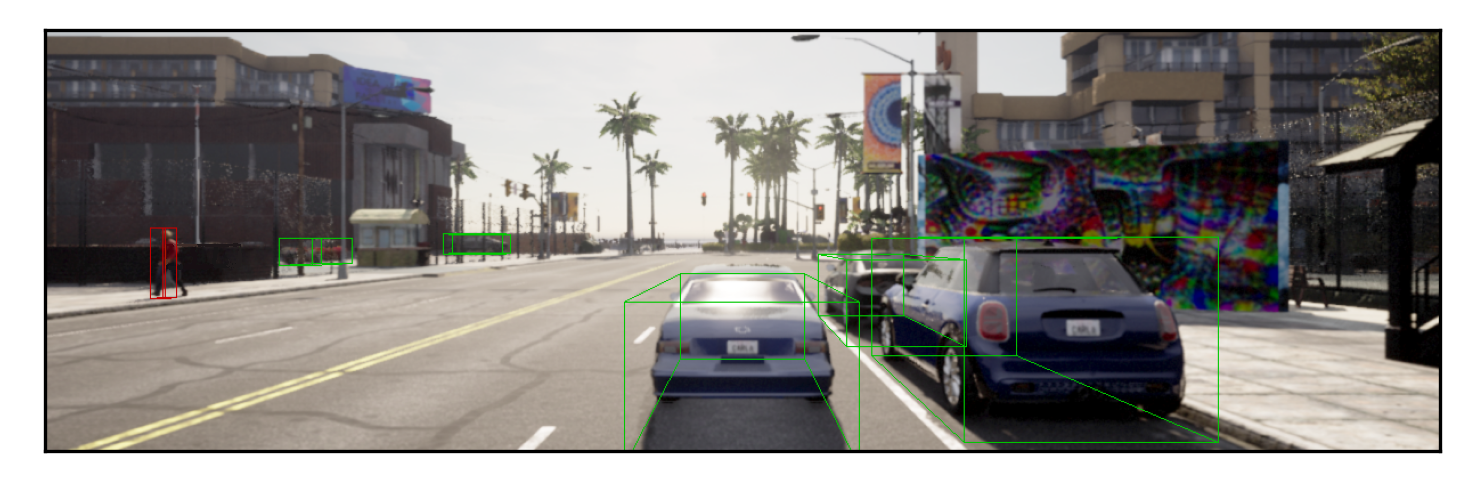}
\end{subfigure} &
\begin{subfigure}[t]{\colb}
    \centering
    \includegraphics[width=\textwidth]{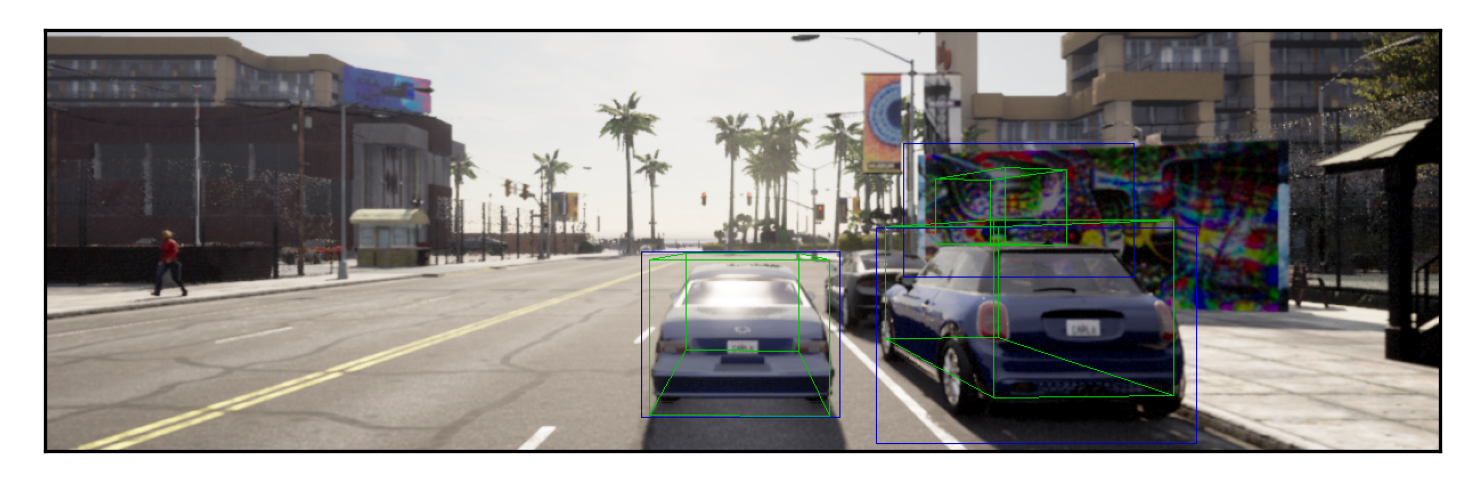}
\end{subfigure}& \rotatebox[origin=r]{-90}{\raggedleft \scriptsize  Stereo R-CNN}
\end{tabular}
\vspace{-0.4cm}\caption[short]{\small Examples of different attack scenarios and tasks. Best viewed in digital, zooming in.
}\label{fig:examples}
\end{figure}

Adversarial examples represent a serious threat in several application domains. They can induce a convolutional neural network (CNN) to produce a wrong output by modifying the input with malicious perturbations that are typically imperceptible to humans. In the last years, an impressive amount of literature has been produced on this topic.

Physical adversarial examples are special types of attacks carried out by means of physical objects, which are crafted to induce an adversarial behavior when captured by a camera whose images are processed by a CNN. They are even more subtle than those acting on the digital representation of images, as they can fool a CNN without directly accessing the vision system, but acting on the external environment. This type of attacks represents a serious threat for safety-critical systems relying on vision-based perception (e.g., self-driving cars), and a general countermeasure has yet to be found; hence, an extensive evaluation of the adversarial robustness of CNNs is a crucial step for increasing the security of such systems.

In spite of the high relevance of such a step, no much attention has been devoted in the literature to the development of datasets or benchmarks for a systematic evaluation of the adversarial robustness of CNNs against physical adversarial attacks, nor for assessing the performance of adversarial defense methods in a unified framework, in particular for autonomous driving scenarios. This might be due to the practical difficulties posed by such an evaluation in the autonomous driving domain, which could result incomplete and/or dangerous. However, the rise of high-definition simulators is paving the way for fully-controllable, photo-realistic driving scenarios that allow for an extensive evaluation of potentially dangerous situations.

This lack in the literature motivates the quest for the design of datasets and benchmarks for the evaluation of the adversarial robustness of CNNs and the performance of defense methods in the physical world. To this end, this paper presents \NAME ~ (\textbf{Ge}neration of datasets for \textbf{A}dversarial \textbf{R}obustness evaluation with CARLA), a tool built on top of the Python API of the CARLA simulator \cite{2017arXiv171103938D}, which allows constructing photo-realistic synthetic datasets for four vision tasks, namely semantic segmentation, 2D and 3D stereo object detection, and monocular depth estimation. 

Figure \ref{fig:examples} illustrates some representative driving scenarios considered in this work. 
Given a urban scenario where an adversarial patch might be placed (e.g., on a billboard or a truck) and, optionally, an adversarial patch, \NAME~attaches the patch on the selected surface and iteratively places the vehicle and the attached camera around the scene, collecting high-definition RGB images, the ground-truth labels, and additional information on the camera intrinsic and extrinsic matrices, as well as the position of the billboard in the scene.

If the patch is not available, it can be generated by first collecting a dataset around a certain billboard with no patch attached, and then running the optimization algorithm proposed in \cite{9706854}. The resulting patched dataset can be used to evaluate the performance of different defense mechanisms, or to evaluate the adversarial robustness of a target CNN. Several datasets, including different attack scenarios, can be collected and used for a more systematic evaluation. Together with the generation tool, this paper presents an extensive comparison of the performance of a selection of adversarial defense and detection methods on several datasets produced with \NAME. Such datasets are available at the project main webpage: \url{http://carlagear.retis.santannapisa.it}.

\NAME~is licensed under MIT License and is hence free and open to use (similarly to CARLA). Ethically, this work raises security and safety concerns about the deployment of safety-critical systems based on vision. While it is true that it might be used with malicious intent, we are hoping to provide a useful tool to the community to help shed a light on the performance of neural networks and defense strategies against adversarial attacks in real-world, fully-controllable scenarios. \\

\pagebreak
The contributions of this paper can be summarized as follows:
\begin{compactitem}
    \item It presents \NAME, the first tool for automatic generation of datasets for (i) evaluating the adversarial robustness of CNNs in the physical world and (ii) comparing the performance of different adversarial defense/detection methods in the context of autonomous driving for four different computer vision tasks: semantic segmentation, monocular depth estimation, 2d object detection, stereo 3d object detection. The tool is available at \url{https://github.com/retis-ai/CARLA-GeAR}.

    \item It presents an extensive comparison of a selection of adversarial defense and detection methods on a set of datasets created with \NAME, suggesting that such a tool can be used to generate benchmarks for a standardized comparison of defense methods. The datasets and corresponding patch generation and evaluation code are available at \url{http://carlagear.retis.santannapisa.it}.
\end{compactitem}

The remainder of this paper is organized as follows: Section \ref{s:related} reviews the related literature, Section \ref{s:proposed} presents \NAME~and the generation of adversarial patches, Section \ref{s:exp} reports the experimental results, and Section \ref{s:conclusions} discusses the limitations of the approach, future work, and states the conclusions.

\section{Related Work} \label{s:related}

Since the discovery of the phenomenon of adversarial examples \cite{biggio2013evasion}, \cite{DBLP:journals/corr/SzegedyZSBEGF13}, a huge amount of work has been produced around them.\footnote{\url{https://nicholas.carlini.com/writing/2019/all-adversarial-example-papers.html}} Adversarial attacks can be coarsely divided in two categories: digital attacks, for which the attacker has complete control of the digital representation of the image, and physical-attacks \cite{kurakin_adversarial_2017}, mainly realized by patches \cite{brown_adversarial_2018} that can be printed and placed in the real world.

Digital attacks are typically composed of image-specific, norm-limited, full-image perturbations and received most of the attention by researchers because of their effectiveness. However, the scope of these attacks is limited to those systems that allow the upload of a user-defined image. On the other hand, physical adversarial attacks require more complex optimization to make the perturbation robust to viewpoint and illumination changes \cite{pmlr-v80-athalye18b}. Furthermore, they are image-agnostic perturbations \cite{moosavi2017universal}
that can be printed and placed in the physical world to fool CNNs operating in the wild. This fact might be a serious concern for safety-critical systems relying on CNN-based vision systems such as autonomous driving. Therefore, a crucial step for a safe and secure deployment of such systems in the real world is an extensive evaluation of their robustness against this kind of attacks.

Despite the large success of adversarial attacks, only a few works proposed datasets and benchmarks for a systematic evaluation of the robustness of models, and most of them is focused on digital adversarial examples \cite{dong2020benchmarking}, \cite{croce2020robustbench}, \cite{mnistc}, \cite{biggio2022patches} and none of them considers the autonomous driving context. In \cite{jefferson_robust_2020}, \cite{2022arXiv220101850R}, the authors sytematically evaluate the effectiveness of real-world patches. To the author's best records, APRICOT \cite{braunegg2020apricot} is the only publicly available dataset that includes physical-world adversarial patches. However, it can only be used to test 2D object detection models. Furthermore, the dataset does not include non-adversarial images and the patches are not always effective due to patch bending or extreme view angles. Conversely, \NAME~is available for four different vision tasks for autonomous driving and, being built on the CARLA Python API, there is full control of the different situations.

The adversarial patches included in the static dataset provided were generated using the adversarial pipeline proposed in \cite{9706854} to perform untargeted white-box attacks. 

Among all the adversarial defenses and detection methods, we restricted the search to those methods designed to detect/mask adversarial patches that are easily extendable to the physical world case and to different tasks. In particular, we chose Z-Mask \cite{2022arXiv220307341R}, FPDA \cite{2022arXiv220101850R}, LGS \cite{naseer_local_2019}, HyperNeuron \cite{hyperneuron}. Additional details are presented in the supplementary material\footnote{Supplementary material are available upon request to the authors.}.

\pagebreak 
\section{Proposed Dataset Generation} \label{s:proposed}

The proposed pipeline is designed to generate datasets for four different computer vision tasks involved in autonomous driving perception. Each dataset is intended to evaluate the adversarial robustness of a CNN and/or the performance of a defense method in a situation in which a physical adversarial attack might be present (chosen from a library of attack situations). For each situation, the ego vehicle with its cameras is iteratively spawned at different distances from the attackable surface, together with the random non-playing characters (NPCs, i.e., vehicles and pedestrians). The dataset is composed of the so-collected RGB images and the corresponding ground-truth annotations, which are different for each task. The supplementary material includes in-depth explanations of the generation and collection process.

\subsection{Attack situations} \label{ss:situations}
This work considers real-world adversarial attacks based on patches. The tool considers two different types of situations: (i) a patch (or two in the case of multi-patch attacks) on a billboard on the side of the road, and (ii) a patch on the back of a truck placed in front of the camera. 
Each billboard situation is specified in a \texttt{yml} file as the fixed position of the billboard in the map. The same file specifies the spawn positioning limits of the ego vehicle and the NPCs relative to the billboard. The spawning is then randomized within these limits. In this way, it is possible to generate different views of the same potentially dangerous scene.
While the billboards have fixed positions, the truck is randomly spawned in the map, and the ego vehicle is spawned behind it at different randomized distances.

\subsection{Configuration files} \label{ss:configuration}

Three configuration files are required: (i) \texttt{collection\_config.yml}, which is the main collection configuration file, (ii) the
\texttt{billboard\_config.yml} file, which includes information about the billboard positioning and the limits for the spawning area of the ego and NPC vehicles, and (iii) the
\texttt{simulation\_config.py} file, which sets the seed and other simulation parameters (number of NPCs in a scene and other Unreal utilities). The role of these configuration files is explained in detail in the supplementary material.

The pipeline described in this section generates single dataset splits (train, validation, test, and so on) starting from the main configuration file \texttt{collection\_config.yml}, that specifies:
\begin{compactitem}
    \item The target task (one between semantic segmentation, 2D object detection, stereo 3D object detection, monocular depth estimation). This directly specifies the folder structure and the ground truth annotation types required;
    \item The CARLA town. Any CARLA town would work, but the one used throughout this paper is Town10HD, since it is the town with the most photo-realistic meshes in CARLA;
    \item The dataset root folder and split; 
    \item The desired scene. As explained in Section \ref{ss:situations}, this configures the adversarial surface positioning, the ego vehicle, and NPCs spawning.
    \item The desired patch to be uploaded and additional info on it. The patch path, if specified, is used to render the patch on the surface selected. If it is not specified, no patch is rendered. This possibility is better detailed in Section \ref{ss:patch_generation}.
\end{compactitem} 

The \texttt{billboard\_config.yml} and \texttt{simulation\_config.py} configuration files are then read to define the attack situation and simulation settings.


\subsection{Data generation algorithm} \label{ss:pipeline}

\begin{figure}[ht]
\begin{center}
\includegraphics[width=\textwidth]{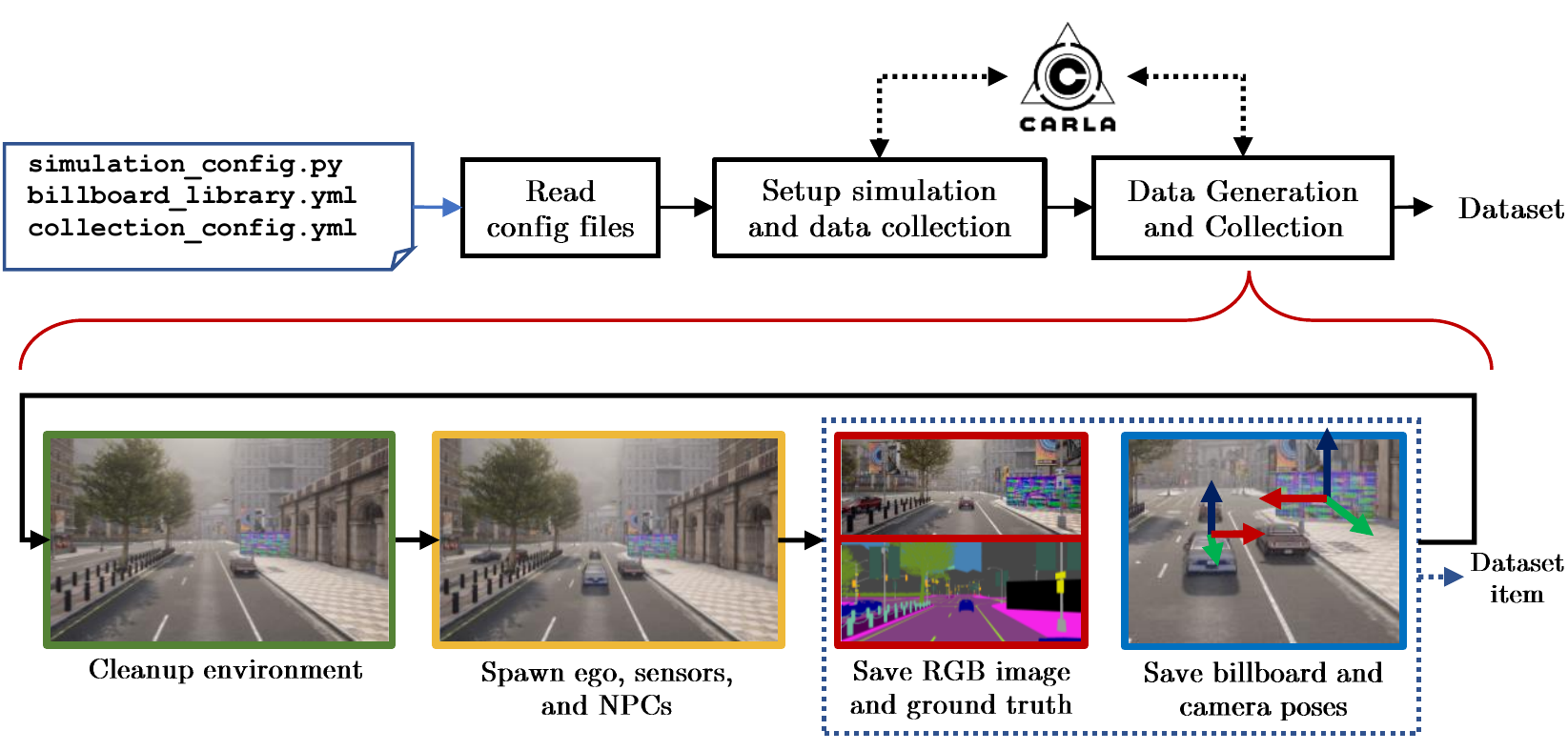}
\caption{Complete generation flow (top) and detail of the data generation and collection phase.}
\label{f:data_gen}
\end{center}
\vskip -0.2in
\end{figure}

Figure \ref{f:data_gen} shows the data generation and collection pipeline, which is discussed next.

\textbf{Read config files} is a set of parsers that extract information from the configuration files presented in Section \ref{ss:configuration}.

\textbf{Setup simulation and data collection} sets up the client-server communication with CARLA through its Python API and writes a few necessary settings. It then loads the specified town and spawns the selected billboards. It also sets up the camera types required for the specific task, the seed for repeatability, and generates the folder tree to properly store images and annotations.

\textbf{Data Generation and Collection} is the core of the pipeline illustrated in Figure \ref{f:data_gen}. The dataset is constructed by iterating three steps: (i) cleanup of any additional vehicle/pedestrian, (ii) spawn of the ego vehicle, its sensors, and randomized NPCs following the spawning limits of the specific situation, (iii) save the RGB image, the ground truth (this task-specific operation is detailed in Section \ref{ss:gt_gen}), the billboard, and the camera poses. Additional details are provided in the supplementary material.

\subsection{Dataset structure and annotations} \label{ss:gt_gen}

RGB images are always saved as uint8 with different resolutions, while each task has different ground-truth annotations and folder structure: semantic segmentation datasets follow the CityScapes \cite{DBLP:conf/cvpr/CordtsORREBFRS16} format, 2D object detection uses the COCO format \cite{lin2014microsoft}, stereo 3D object detection the Kitti Stereo Object Detection format\cite{KITTI}, and monocular depth estimation the Kitti Depth format. 
This choice is motivated by the fact that the same original metrics, evaluation procedures and deep learning framework data loaders can be used to evaluate these custom datasets. 

CARLA has special semantic segmentation and depth camera sensors that allow immediate ground-truth computation. It can also compute all the 3D bounding boxes in the simulated world. However, the 2D and 3D bounding boxes that are visible in each scene must be computed heuristically (details are reported in the supplementary material).

Additionally, \NAME~collects all the billboard positions together with the camera extrinsic and intrisic matrices. This allows knowing the precise pose of the attackable surface in the scene, which is crucial to make an accurate digital patch placement (see details in Section \ref{ss:patch_generation}) and obtain the ground truth for evaluating the accuracy of masking defense methods against adversarial patches.


\subsection{Patch generation} \label{ss:patch_generation}

As described in Section \ref{ss:situations}, the \texttt{collection\_config.yml} file specifies the path of the patch that must be uploaded and rendered on the selected billboard. If the file path is not defined, the tool does not render anything on the billboard. This is useful to have an additional test split to check the performance of a CNN or a defense method in a non-adversarial case. 
Furthermore, these non-adversarial dataset splits can be used to train adversarial patches for that specific situation. 
We follow the \emph{scene-specific} attack method proposed in \cite{9706854}, which requires camera and billboard poses to accurately reproject the digital patch to maintain differentiability and the possibility to perform white-box attacks. Implementation details are reported in Section \ref{ss:setup}.
\pagebreak
\section{Experimental Results} \label{s:exp}
This section presents the results of preliminary tests that were performed to set the simulation configuration used for the generation of datasets (Section \ref{ss:ablation}). Furthermore, in Section \ref{ss:defense} the datasets generated are used to compare a selection of state-of-the-art defenses and detection methods for each task. The experimental setup is briefly listed in Section \ref{ss:setup} and better detailed in the supplementary material, together with additional experiments and illustrations from the datasets.

\subsection{Experimental setup} \label{ss:setup}
The CARLA simulator version 0.9.13 was used (UnrealEngine 4.26 on Ubuntu 18.04) on a machine equipped with an Intel i7-4790K CPU @ 4.00GHz × 8 and an NVidia GTX 1080Ti GPU. The adversarial patch optimizations were performed on an NVidia Tesla A100 GPU using PyTorch. 

\textbf{The CNNs} used in this paper were DDRNet23Slim \cite{ddrnet_paper} and BiSeNetXception39 \cite{bisenet_paper} for semantic segmentation, Faster R-CNN \cite{7485869} and RetinaNet \cite{lin2017focal} for 2D object detection, GLPDepth \cite{kim2022global} and AdaBins \cite{Bhat_2021_CVPR} for monocular depth estimation and Stereo R-CNN \cite{li_stereo_2019} for stereo 3D object detection.

\textbf{The defenses} used for the comparison are FPDA \cite{2022arXiv220101850R}, Z-mask \cite{2022arXiv220307341R} and HyperNeuron \cite{hyperneuron} for adversarial detection, and Z-mask \cite{2022arXiv220307341R} and LGS \cite{naseer_local_2019} for adversarial defense.

\textbf{The comparison metrics} used in this paper are (i) mIoU for semantic segmentation, (ii) the COCO mAP for 2D object detection, (iii) root mean square error (RMSE) in meters for monocular depth estimation, and (iv) kitti AP for ``moderate" label difficulty for stereo 3D object detection. Please note that additional metrics are reported in the supplementary materials, but, nonetheless, these metrics do not entirely reflect the effectiveness of a real-world attack or defense: the labels produced by CARLA are often too fine-grained and occlusions might occur. Hence, to properly reflect the performance of the analyzed scenarios there is a strong need for specialized metrics designed for this purpose. This is left as a future work in perspective of the construction of a benchmark--- a discussion of this issue is presented in Section \ref{s:conclusions}.
The performance of adversarial detection algorithms is evaluated with the area under the receiver operating characteristic (AUROC). Each ROC is computed on two ``twin" datasets, one including the adversarial patches and the other containing the same images, but without any patch.

\textbf{Life-cycle of a patch}.
To evaluate the effects of real-world adversarial patch attacks in CARLA it is necessary to follow a few steps. Given a certain task and an attack scenario, it is possible to collect a training dataset split to digitally optimize the patch to attack a specific CNN using the optimization algorithm described in \cite{9706854}. The same training set can be used to craft patches for different networks for the same task. Once the patch is optimized, it can be imported in CARLA and applied to the billboard to generate the test set. In this way, it is rendered realistically as the other objects in the scene and it constitutes a ``virtual" real-world adversarial object. Different patches (for different networks) are used to generate different test sets. However, each test set shares the same seed. This allows the generation of datasets with exactly the same images, except for the selected patch, hence providing a more fair comparison. Furthermore, each adversarial patch is tested against a control test set that includes a random patch or no patch.

The datasets used in Section \ref{ss:defense} include 10 attack situations, namely 9 billboard-based attacks (billboard01 to billboard09, of which billboard05 to billboard07 are double billboard for double patch attacks) and 1 truck-based attack.


\subsection{Ablation studies} \label{ss:ablation}

This section reports the results of ablation studies performed in the early stages of the experimentation to set the parameters used for the data generation. The objective was to find a good trade-off between the CNN performance and the attack effectiveness while maintaining reasonable computation times.
Inspired by a previous work \cite{2022arXiv220101850R}, we decided to use the same billboard configuration that allows the application of a 3.7m$\times$7.4m patch with 150$\times$300 pixels.
The CNNs used for these studies are Faster R-CNN for 2D object detection, GLPDepth for depth estimation, and DDRNet for semantic segmentation.

\paragraph{Computation time.}

\begin{wrapfigure}{r}{0.45\textwidth}
    \centering
    \includegraphics[width=\textwidth]{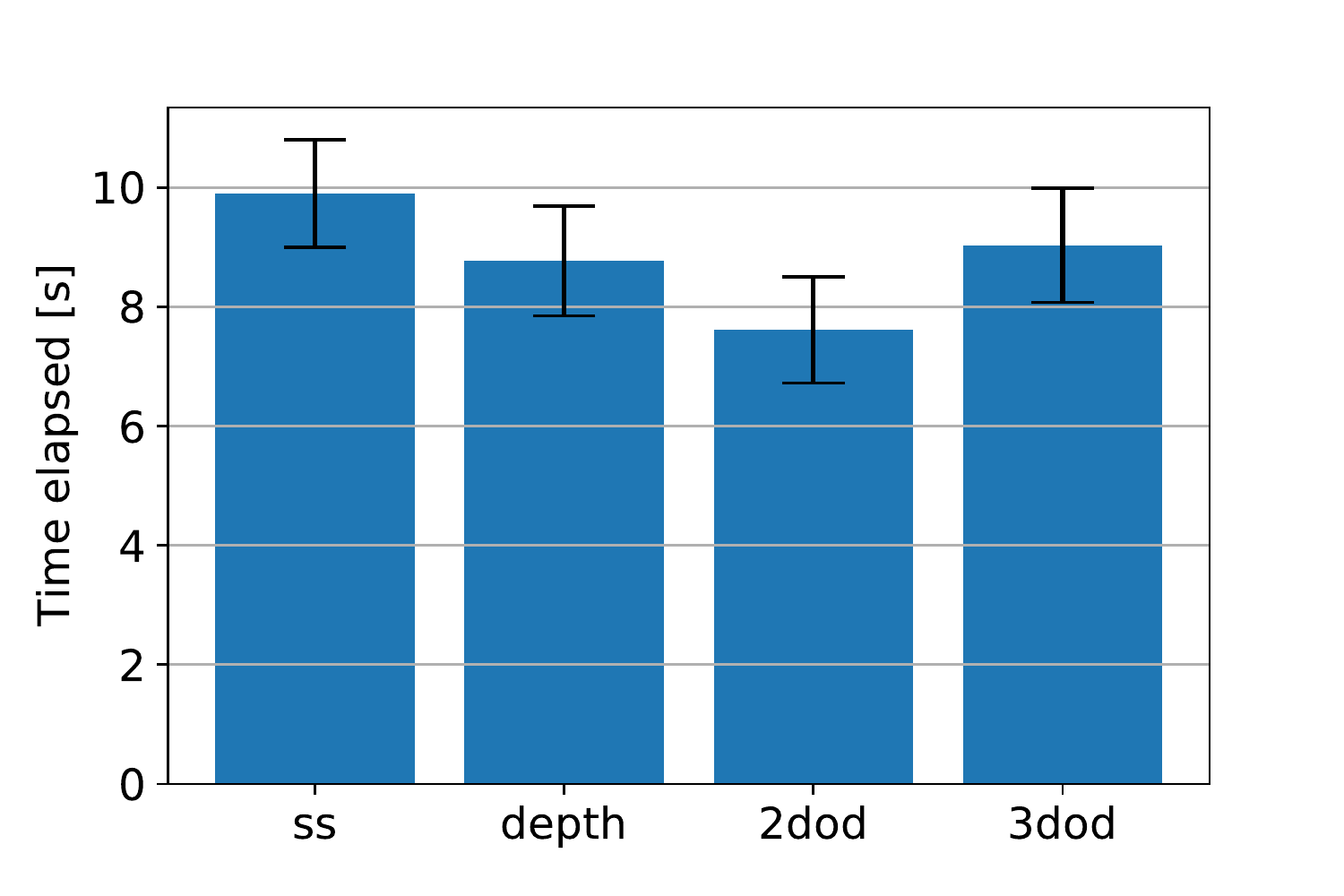}
    \vskip -0.1 in
\caption{\small Time required for the generation of a dataset item for each task. The error bar represents the st. dev.}\label{fig:times}
\vskip -0.1in
\end{wrapfigure}

The creation of such datasets is a time-consuming task. In this section we show the average time required to generate and save an RGB image and ground-truth annotation for each task.
From the results reported in Figure \ref{fig:times}, it is clear that the most time-consuming task is semantic segmentation. This is attributed to the fact that segmented images (and ground-truth labels) have the highest resolution, following the Cityscapes format. Then, stereo 3D object detection and depth estimation follow, since the first saves pairs of RGB images and the latter saves an RGB and a depth image. 2D object detection is the least expensive task for generation, since it has to save the RGB image only, while the annotation is a json file. The timing data have been collected during the generation of the same 50-sample test set, changing the seed ten times. Hence, the generation of a 50-sample dataset takes roughly eight minutes on average.

\begin{figure}[ht]
\begin{subfigure}{0.2\textwidth}
    \raggedleft
    \includegraphics[width=\textwidth]{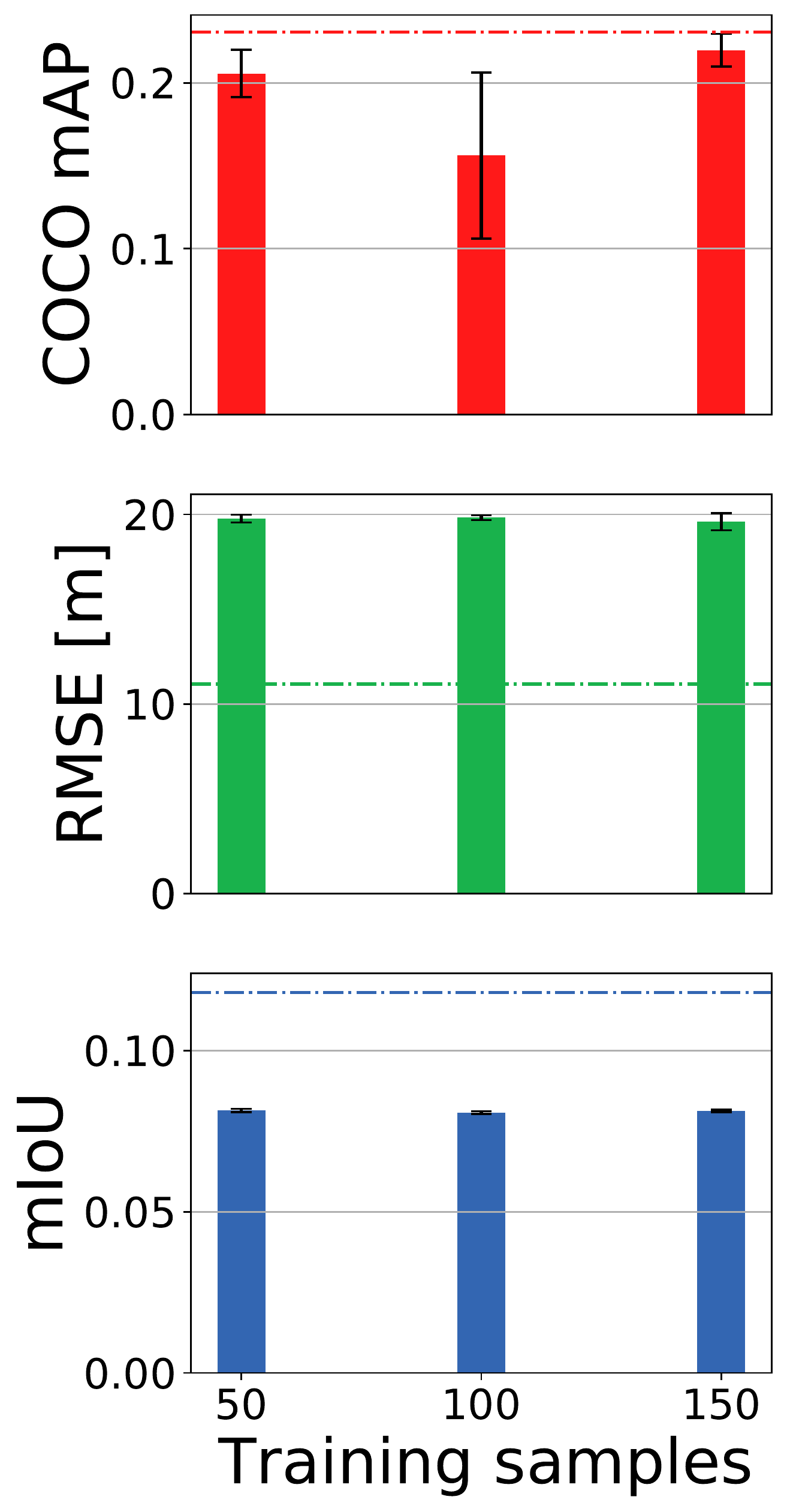}
    \caption{}\label{f:ablation_samples}
\end{subfigure}%
\begin{subfigure}{0.9\textwidth}
    \vskip -0.2 in
    \raggedleft
    \includegraphics[width=\textwidth]{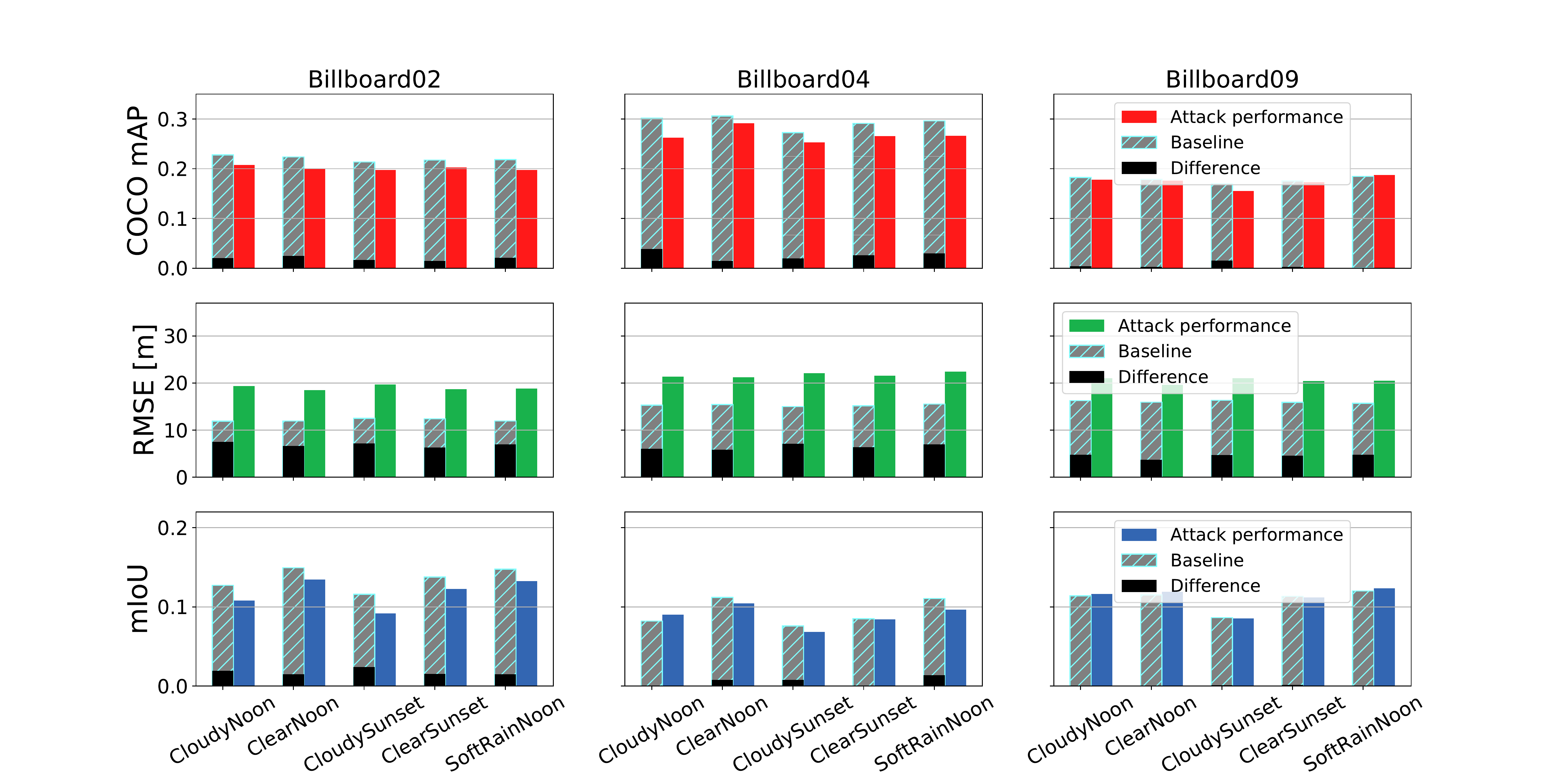}
    \caption{}\label{f:weather}
\end{subfigure}
\caption{\small (a) Performance of the attack for each task as a function of the number of samples used for the optimization of the image-agnostic patch. The experiment was performed on billboard02 ten times, changing the seed for each experiment. The dash-dotted line represents the baseline performance (random patch) whereas the error bar represents the standard deviation. (b) Effect of the different weather presets of CARLA on the baseline performance (random patch) against the attack effectiveness for each task. The experiments were performed for three different attack situations with different camera-sunlight relative orientations.}
\label{f:ablation_weather}
\vskip -0.1in
\end{figure}

\paragraph{Number of samples for the optimization.}
An important parameter that must be set is the number of samples required to run the optimizations properly. Too few samples might not be enough to obtain effective attacks, while too many samples slow the generation and optimization processes down. Figure \ref{f:ablation_samples} shows the attack performance with respect to the baseline CNN (obtained with a random patch) as a function of the number of samples for each task. The results were averaged over ten different optimizations, changing the seed each time.
Since optimizing a patch with 100 samples leads to better results overall, we set the training set dimension to 100 images in the generation stage.

\paragraph{Weather conditions.}
The CARLA simulator allows full control of the weather parameters, ranging from the time of day (i.e., the elevation angle of the sun) to the amount of rain and wetness of the road, from the fog density to the Rayleigh scattering. Given the large number of different parameters to evaluate, which would lead to a combinatorial explosion in the number of settings, we decided to limit the study to the weather presets available in CARLA, ignoring those that lie outside the distribution of the datasets used to train the CNNs under test (i.e., no heavy rain, puddles, night, or fog). The tested weather presets were \texttt{CloudyNoon}, \texttt{ClearNoon}, \texttt{CloudySunset}, \texttt{ClearSunset}, \texttt{SoftRainNoon}.
Nevertheless, note that in \NAME~ the weather parameters are fully configurable. 

We decided to restrict the search to three attack situations with different billboard orientation (and, hence, camera orientation) with respect to the sunlight position: billboard02 has the sun behind (shining directly on the billboard), billboard04 has the sun on the side, and billboard09 has the sun in front (behind the billboard). 
The results in Figure \ref{f:weather} show that, for each task tested, the baseline performance (i.e., applying a random real-world patch) when using presets \texttt{CloudySunset} and \texttt{ClearSunset} is almost always slightly worse. The attack performance resulted not to be particularly influenced by the weather conditions. Hence, we used the \texttt{ClearNoon} weather for the following experiments, since it performs slightly better overall.

\begin{wraptable}{r}{0.48\textwidth}\resizebox{\textwidth}{!}{\begin{tabular}{ll|ll|ll|ll|l|}\cline{3-9}                                                   &     & \multicolumn{2}{l|}{SS}               & \multicolumn{2}{l|}{2DOD}              & \multicolumn{2}{l|}{Depth}              & 3DOD        \\ \hline\multicolumn{1}{|l|}{}                        & Def & \multicolumn{1}{l|}{\tiny DDRNet} & \tiny BiseNet & \multicolumn{1}{l|}{\tiny FRCNN} & \tiny RetinaNet & \multicolumn{1}{l|}{\tiny AdaBins} & \tiny GLPDepth & \tiny Stereo RCNN \\ \hline\multicolumn{1}{|l|}{\multirow{3}{*}{\rotatebox[origin=c]{90}{\tiny Billboard01}}} & ZM  & \multicolumn{1}{l|}{0.99}       &0.95         & \multicolumn{1}{l|}{0.50}      &0.49           & \multicolumn{1}{l|}{0.27}        &0.00          &0.82             \\ \cline{2-9} \multicolumn{1}{|l|}{}                             & FPDA & \multicolumn{1}{l|}{1.00}       &0.97         & \multicolumn{1}{l|}{0.53}      &0.49           & \multicolumn{1}{l|}{0.20}        &0.49          &0.50             \\ \cline{2-9} \multicolumn{1}{|l|}{}                             & HN & \multicolumn{1}{l|}{1.00}       &1.00         & \multicolumn{1}{l|}{0.52}      &0.53           & \multicolumn{1}{l|}{0.09}        &0.12          &0.89             \\ \hline\multicolumn{1}{|l|}{\multirow{3}{*}{\rotatebox[origin=c]{90}{\tiny Billboard02}}} & ZM  & \multicolumn{1}{l|}{1.00}       &1.00         & \multicolumn{1}{l|}{1.00}      &1.00           & \multicolumn{1}{l|}{0.89}        &0.26          &0.92             \\ \cline{2-9} \multicolumn{1}{|l|}{}                             & FPDA & \multicolumn{1}{l|}{1.00}       &1.00         & \multicolumn{1}{l|}{0.97}      &0.66           & \multicolumn{1}{l|}{0.90}        &0.00          &0.50             \\ \cline{2-9} \multicolumn{1}{|l|}{}                             & HN & \multicolumn{1}{l|}{1.00}       &1.00         & \multicolumn{1}{l|}{0.97}      &0.78           & \multicolumn{1}{l|}{0.98}        &0.51          &0.96             \\ \hline\multicolumn{1}{|l|}{\multirow{3}{*}{\rotatebox[origin=c]{90}{\tiny Billboard03}}} & ZM  & \multicolumn{1}{l|}{0.98}       &0.90         & \multicolumn{1}{l|}{0.99}      &0.99           & \multicolumn{1}{l|}{0.99}        &0.33          &0.74             \\ \cline{2-9} \multicolumn{1}{|l|}{}                             & FPDA & \multicolumn{1}{l|}{0.86}       &0.92         & \multicolumn{1}{l|}{0.64}      &0.85           & \multicolumn{1}{l|}{0.89}        &0.43          &0.50             \\ \cline{2-9} \multicolumn{1}{|l|}{}                             & HN & \multicolumn{1}{l|}{0.95}       &0.87         & \multicolumn{1}{l|}{0.88}      &0.82           & \multicolumn{1}{l|}{0.97}        &0.23          &0.75             \\ \hline\multicolumn{1}{|l|}{\multirow{3}{*}{\rotatebox[origin=c]{90}{\tiny Billboard04}}} & ZM  & \multicolumn{1}{l|}{1.00}       &1.00         & \multicolumn{1}{l|}{0.95}      &0.96           & \multicolumn{1}{l|}{0.97}        &0.39          &0.97             \\ \cline{2-9} \multicolumn{1}{|l|}{}                             & FPDA & \multicolumn{1}{l|}{0.99}       &1.00         & \multicolumn{1}{l|}{0.58}      &0.56           & \multicolumn{1}{l|}{0.92}        &0.49          &0.50             \\ \cline{2-9} \multicolumn{1}{|l|}{}                             & HN & \multicolumn{1}{l|}{1.00}       &1.00         & \multicolumn{1}{l|}{0.70}      &0.66           & \multicolumn{1}{l|}{0.97}        &0.40          &0.95             \\ \hline\multicolumn{1}{|l|}{\multirow{3}{*}{\rotatebox[origin=c]{90}{\tiny Billboard05}}} & ZM  & \multicolumn{1}{l|}{1.00}       &1.00         & \multicolumn{1}{l|}{0.86}      &0.90           & \multicolumn{1}{l|}{0.31}        &0.05          &0.92             \\ \cline{2-9} \multicolumn{1}{|l|}{}                             & FPDA & \multicolumn{1}{l|}{0.99}       &1.00         & \multicolumn{1}{l|}{0.62}      &0.64           & \multicolumn{1}{l|}{0.30}        &0.50          &0.78             \\ \cline{2-9} \multicolumn{1}{|l|}{}                             & HN & \multicolumn{1}{l|}{0.99}       &1.00         & \multicolumn{1}{l|}{0.74}      &0.72           & \multicolumn{1}{l|}{0.27}        &0.39          &0.95             \\ \hline\multicolumn{1}{|l|}{\multirow{3}{*}{\rotatebox[origin=c]{90}{\tiny Billboard06}}} & ZM  & \multicolumn{1}{l|}{1.00}       &1.00         & \multicolumn{1}{l|}{1.00}      &1.00           & \multicolumn{1}{l|}{0.96}        &0.59          &0.98             \\ \cline{2-9} \multicolumn{1}{|l|}{}                             & FPDA & \multicolumn{1}{l|}{1.00}       &1.00         & \multicolumn{1}{l|}{0.87}      &0.61           & \multicolumn{1}{l|}{0.86}        &0.07          &0.50             \\ \cline{2-9} \multicolumn{1}{|l|}{}                             & HN & \multicolumn{1}{l|}{1.00}       &1.00         & \multicolumn{1}{l|}{0.85}      &0.85           & \multicolumn{1}{l|}{0.88}        &0.94          &0.99             \\ \hline\multicolumn{1}{|l|}{\multirow{3}{*}{\rotatebox[origin=c]{90}{\tiny Billboard07}}} & ZM  & \multicolumn{1}{l|}{1.00}       &0.60         & \multicolumn{1}{l|}{1.00}      &1.00           & \multicolumn{1}{l|}{1.00}        &0.61          &0.93             \\ \cline{2-9} \multicolumn{1}{|l|}{}                             & FPDA & \multicolumn{1}{l|}{1.00}       &1.00         & \multicolumn{1}{l|}{0.89}      &0.60           & \multicolumn{1}{l|}{0.97}        &0.49          &0.50             \\ \cline{2-9} \multicolumn{1}{|l|}{}                             & HN & \multicolumn{1}{l|}{1.00}       &1.00         & \multicolumn{1}{l|}{0.96}      &0.73           & \multicolumn{1}{l|}{0.98}        &0.27          &0.92             \\ \hline\multicolumn{1}{|l|}{\multirow{3}{*}{\rotatebox[origin=c]{90}{\tiny Billboard08}}} & ZM  & \multicolumn{1}{l|}{1.00}       &0.97         & \multicolumn{1}{l|}{0.66}      &0.67           & \multicolumn{1}{l|}{0.29}        &0.00          &0.98             \\ \cline{2-9} \multicolumn{1}{|l|}{}                             & FPDA & \multicolumn{1}{l|}{0.87}       &1.00         & \multicolumn{1}{l|}{0.58}      &0.55           & \multicolumn{1}{l|}{0.23}        &0.50          &0.50             \\ \cline{2-9} \multicolumn{1}{|l|}{}                             & HN & \multicolumn{1}{l|}{0.93}       &1.00         & \multicolumn{1}{l|}{0.59}      &0.67           & \multicolumn{1}{l|}{0.21}        &0.10          &0.99             \\ \hline\multicolumn{1}{|l|}{\multirow{3}{*}{\rotatebox[origin=c]{90}{\tiny Billboard09}}} & ZM  & \multicolumn{1}{l|}{0.68}       &0.65         & \multicolumn{1}{l|}{0.84}      &0.86           & \multicolumn{1}{l|}{0.91}        &0.48          &0.59             \\ \cline{2-9} \multicolumn{1}{|l|}{}                             & FPDA & \multicolumn{1}{l|}{0.62}       &1.00         & \multicolumn{1}{l|}{0.58}      &0.46           & \multicolumn{1}{l|}{0.90}        &0.51          &0.50             \\ \cline{2-9} \multicolumn{1}{|l|}{}                             & HN & \multicolumn{1}{l|}{0.61}       &1.00         & \multicolumn{1}{l|}{0.76}      &0.51           & \multicolumn{1}{l|}{0.99}        &0.51          &0.65             \\ \hline\multicolumn{1}{|l|}{\multirow{3}{*}{\rotatebox[origin=c]{90}{\tiny Truck}}} & ZM  & \multicolumn{1}{l|}{0.87}       &0.68         & \multicolumn{1}{l|}{0.58}      &0.56           & \multicolumn{1}{l|}{0.60}        &0.42          &0.63             \\ \cline{2-9} \multicolumn{1}{|l|}{}                             & FPDA & \multicolumn{1}{l|}{0.87}       &0.72         & \multicolumn{1}{l|}{0.52}      &0.54           & \multicolumn{1}{l|}{0.52}        &0.46          &0.50             \\ \cline{2-9} \multicolumn{1}{|l|}{}                             & HN & \multicolumn{1}{l|}{0.86}       &0.77         & \multicolumn{1}{l|}{0.53}      &0.52           & \multicolumn{1}{l|}{0.58}        &0.43          &0.62             \\ \hline\end{tabular}}
\vspace{-0.12in}
\caption{\small Detection performance comparison (AUROC) for Z-Mask, FPDA, and HyperNeuron for each attack situation and each task.}\label{t:detection}\end{wraptable}



\subsection{Defense comparison} \label{ss:defense}

This section presents an extensive comparison of different defense methods on the dataset generated for each of the four tasks considered. 

Table \ref{t:defenses} reports the performance of each defense method considered for each task. Notably, there are a few attack situations that do not induce a large adversarial effect (e.g., Billboard01, 02, 03, 05, 09 and Truck on DDRNet), where the defense is actually harming the performance of the network. Furthermore, it is worth noting how the mAP metric for 2D object detection indicates a consistent low adversarial effect and bad defense performance, while in reality the effect of the adversarial patch is always present (see Figure \ref{fig:examples} for an example) and the defense is masking the patch with fair accuracy. 
This observation highlights the need for a more accurate metric to evaluate the effect of these patches (and corresponding defense) and constitutes one of the limitations to the construction of a benchmark from these situations.
Please consult the supplementary material for the results with additional metrics, such as the IoU of the patch mask, and further illustrations.  

Table \ref{t:detection} compares the performance of Z-Mask, FPDA, and HyperNeuron in terms of AUROC. All the methods present good performance for most of the networks, with Z-Mask being the most consistent detection method overall. 

Despite the issues highlighted above, these results still provide relevant quantitative information about the potential power of \NAME~as a generator of adversarial situations for evaluating the behavior of neural models and/or the defense under test.

\begin{table}[]\resizebox{\textwidth}{!}{\begin{tabular}{ll|llll|llll|llll|ll|}\cline{3-16}                                                    &     & \multicolumn{4}{l|}{SS (mIOU)}                                                                                                                  & \multicolumn{4}{l|}{2DOD (mAP)}                                                                                          & \multicolumn{4}{l|}{Depth (RMSE)}                                                                                        & \multicolumn{2}{l|}{3DOD (mAP)}                  \\ \hline\multicolumn{1}{|l|}{}                         & Def & \multicolumn{2}{l|}{DDRNet}                                                                  & \multicolumn{2}{l|}{BiseNet}                     & \multicolumn{2}{l|}{FRCNN}                                            & \multicolumn{2}{l|}{RetinaNet}                   & \multicolumn{2}{l|}{AdaBins}                                          & \multicolumn{2}{l|}{GLPDepth}                    & \multicolumn{2}{l|}{Stereo RCNN}                 \\ \hline\multicolumn{1}{|l|}{}                              & -   & \multicolumn{1}{l|}{\textbf{33.04}} & \multicolumn{1}{l|}{\cellcolor[HTML]{C0C0C0}\color[HTML]{000000}0.65{\color[HTML]{000000} }} & \multicolumn{1}{l|}{\textbf{28.35}} & \cellcolor[HTML]{C0C0C0}\color[HTML]{000000}-10.20 & \multicolumn{1}{l|}{\textbf{19.89}} & \multicolumn{1}{l|}{\cellcolor[HTML]{C0C0C0}\color[HTML]{000000}-0.59} & \multicolumn{1}{l|}{\textbf{20.12}} & \cellcolor[HTML]{C0C0C0}\color[HTML]{000000}-2.66 & \multicolumn{1}{l|}{\textbf{15.51}} & \multicolumn{1}{l|}{\cellcolor[HTML]{C0C0C0}\color[HTML]{000000}6.00} & \multicolumn{1}{l|}{\textbf{13.52}} & \cellcolor[HTML]{C0C0C0}\color[HTML]{000000}3.22 & \multicolumn{1}{l|}{\textbf{51.43}} & \cellcolor[HTML]{C0C0C0}\color[HTML]{000000}-2.84 \\ \cline{2-16} \multicolumn{1}{|l|}{}                              & ZM  & \multicolumn{1}{l|}{\color[HTML]{0000B0}0.00} & \multicolumn{1}{l|}{\cellcolor[HTML]{C0C0C0}\color[HTML]{B00000}-2.31{\color[HTML]{000000} }} & \multicolumn{1}{l|}{\color[HTML]{B00000}-0.15} & \cellcolor[HTML]{C0C0C0}\color[HTML]{00B000}-5.32 & \multicolumn{1}{l|}{\color[HTML]{00B000}0.02} & \multicolumn{1}{l|}{\cellcolor[HTML]{C0C0C0}\color[HTML]{00B000}-0.49} & \multicolumn{1}{l|}{\color[HTML]{B00000}-0.25} & \cellcolor[HTML]{C0C0C0}\color[HTML]{B00000}-2.95 & \multicolumn{1}{l|}{\color[HTML]{00B000}-0.36} & \multicolumn{1}{l|}{\cellcolor[HTML]{C0C0C0}\color[HTML]{00B000}1.72} & \multicolumn{1}{l|}{\color[HTML]{00B000}-0.10} & \cellcolor[HTML]{C0C0C0}\color[HTML]{0000B0}3.22 & \multicolumn{1}{l|}{\color[HTML]{B00000}-0.45} & \cellcolor[HTML]{C0C0C0}\color[HTML]{00B000}-0.86 \\ \cline{2-16} \multicolumn{1}{|l|}{\multirow{-3}{*}{\rotatebox[origin=c]{90}{\tiny Billboard01}}} & LGS & \multicolumn{1}{l|}{\color[HTML]{B00000}-0.07} & \multicolumn{1}{l|}{\cellcolor[HTML]{C0C0C0}\color[HTML]{B00000}0.45{\color[HTML]{000000} }} & \multicolumn{1}{l|}{\color[HTML]{00B000}0.11} & \cellcolor[HTML]{C0C0C0}\color[HTML]{00B000}-8.63 & \multicolumn{1}{l|}{\color[HTML]{00B000}0.02} & \multicolumn{1}{l|}{\cellcolor[HTML]{C0C0C0}\color[HTML]{B00000}-1.38} & \multicolumn{1}{l|}{\color[HTML]{B00000}-0.25} & \cellcolor[HTML]{C0C0C0}\color[HTML]{B00000}-4.41 & \multicolumn{1}{l|}{\color[HTML]{00B000}-0.36} & \multicolumn{1}{l|}{\cellcolor[HTML]{C0C0C0}\color[HTML]{00B000}5.94} & \multicolumn{1}{l|}{\color[HTML]{00B000}-0.10} & \cellcolor[HTML]{C0C0C0}\color[HTML]{00B000}3.12 & \multicolumn{1}{l|}{\color[HTML]{B00000}-0.45} & \cellcolor[HTML]{C0C0C0}\color[HTML]{B00000}-3.09 \\ \hline\multicolumn{1}{|l|}{}                              & -   & \multicolumn{1}{l|}{\textbf{34.11}} & \multicolumn{1}{l|}{\cellcolor[HTML]{C0C0C0}\color[HTML]{000000}-1.74{\color[HTML]{000000} }} & \multicolumn{1}{l|}{\textbf{29.66}} & \cellcolor[HTML]{C0C0C0}\color[HTML]{000000}-4.32 & \multicolumn{1}{l|}{\textbf{19.66}} & \multicolumn{1}{l|}{\cellcolor[HTML]{C0C0C0}\color[HTML]{000000}-0.93} & \multicolumn{1}{l|}{\textbf{19.65}} & \cellcolor[HTML]{C0C0C0}\color[HTML]{000000}-0.45 & \multicolumn{1}{l|}{\textbf{16.04}} & \multicolumn{1}{l|}{\cellcolor[HTML]{C0C0C0}\color[HTML]{000000}4.93} & \multicolumn{1}{l|}{\textbf{12.33}} & \cellcolor[HTML]{C0C0C0}\color[HTML]{000000}5.61 & \multicolumn{1}{l|}{\textbf{38.29}} & \cellcolor[HTML]{C0C0C0}\color[HTML]{000000}3.70 \\ \cline{2-16} \multicolumn{1}{|l|}{}                              & ZM  & \multicolumn{1}{l|}{\color[HTML]{B00000}-0.04} & \multicolumn{1}{l|}{\cellcolor[HTML]{C0C0C0}\color[HTML]{B00000}-1.76{\color[HTML]{000000} }} & \multicolumn{1}{l|}{\color[HTML]{B00000}-0.21} & \cellcolor[HTML]{C0C0C0}\color[HTML]{00B000}-2.35 & \multicolumn{1}{l|}{\color[HTML]{00B000}0.06} & \multicolumn{1}{l|}{\cellcolor[HTML]{C0C0C0}\color[HTML]{B00000}-1.32} & \multicolumn{1}{l|}{\color[HTML]{B00000}-0.17} & \cellcolor[HTML]{C0C0C0}\color[HTML]{B00000}-1.12 & \multicolumn{1}{l|}{\color[HTML]{00B000}-0.02} & \multicolumn{1}{l|}{\cellcolor[HTML]{C0C0C0}\color[HTML]{00B000}0.07} & \multicolumn{1}{l|}{\color[HTML]{00B000}-0.03} & \cellcolor[HTML]{C0C0C0}\color[HTML]{00B000}3.04 & \multicolumn{1}{l|}{\color[HTML]{B00000}-0.41} & \cellcolor[HTML]{C0C0C0}\color[HTML]{B00000}0.85 \\ \cline{2-16} \multicolumn{1}{|l|}{\multirow{-3}{*}{\rotatebox[origin=c]{90}{\tiny Billboard02}}} & LGS & \multicolumn{1}{l|}{\color[HTML]{B00000}-0.20} & \multicolumn{1}{l|}{\cellcolor[HTML]{C0C0C0}\color[HTML]{B00000}-3.51{\color[HTML]{000000} }} & \multicolumn{1}{l|}{\color[HTML]{00B000}0.07} & \cellcolor[HTML]{C0C0C0}\color[HTML]{B00000}-5.09 & \multicolumn{1}{l|}{\color[HTML]{00B000}0.06} & \multicolumn{1}{l|}{\cellcolor[HTML]{C0C0C0}\color[HTML]{B00000}-2.90} & \multicolumn{1}{l|}{\color[HTML]{B00000}-0.17} & \cellcolor[HTML]{C0C0C0}\color[HTML]{B00000}-2.63 & \multicolumn{1}{l|}{\color[HTML]{00B000}-0.02} & \multicolumn{1}{l|}{\cellcolor[HTML]{C0C0C0}\color[HTML]{00B000}3.54} & \multicolumn{1}{l|}{\color[HTML]{00B000}-0.03} & \cellcolor[HTML]{C0C0C0}\color[HTML]{00B000}4.51 & \multicolumn{1}{l|}{\color[HTML]{B00000}-0.41} & \cellcolor[HTML]{C0C0C0}\color[HTML]{00B000}6.23 \\ \hline\multicolumn{1}{|l|}{}                              & -   & \multicolumn{1}{l|}{\textbf{34.39}} & \multicolumn{1}{l|}{\cellcolor[HTML]{C0C0C0}\color[HTML]{000000}-0.60{\color[HTML]{000000} }} & \multicolumn{1}{l|}{\textbf{28.88}} & \cellcolor[HTML]{C0C0C0}\color[HTML]{000000}-2.81 & \multicolumn{1}{l|}{\textbf{19.30}} & \multicolumn{1}{l|}{\cellcolor[HTML]{C0C0C0}\color[HTML]{000000}-0.89} & \multicolumn{1}{l|}{\textbf{18.92}} & \cellcolor[HTML]{C0C0C0}\color[HTML]{000000}-0.69 & \multicolumn{1}{l|}{\textbf{14.23}} & \multicolumn{1}{l|}{\cellcolor[HTML]{C0C0C0}\color[HTML]{000000}3.22} & \multicolumn{1}{l|}{\textbf{12.05}} & \cellcolor[HTML]{C0C0C0}\color[HTML]{000000}3.33 & \multicolumn{1}{l|}{\textbf{77.54}} & \cellcolor[HTML]{C0C0C0}\color[HTML]{000000}-11.56 \\ \cline{2-16} \multicolumn{1}{|l|}{}                              & ZM  & \multicolumn{1}{l|}{\color[HTML]{0000B0}0.00} & \multicolumn{1}{l|}{\cellcolor[HTML]{C0C0C0}\color[HTML]{0000B0}-0.60{\color[HTML]{000000} }} & \multicolumn{1}{l|}{\color[HTML]{0000B0}0.00} & \cellcolor[HTML]{C0C0C0}\color[HTML]{0000B0}-2.81 & \multicolumn{1}{l|}{\color[HTML]{0000B0}0.00} & \multicolumn{1}{l|}{\cellcolor[HTML]{C0C0C0}\color[HTML]{00B000}-0.77} & \multicolumn{1}{l|}{\color[HTML]{0000B0}0.00} & \cellcolor[HTML]{C0C0C0}\color[HTML]{B00000}-0.70 & \multicolumn{1}{l|}{\color[HTML]{00B000}-0.00} & \multicolumn{1}{l|}{\cellcolor[HTML]{C0C0C0}\color[HTML]{00B000}1.03} & \multicolumn{1}{l|}{\color[HTML]{0000B0}0.00} & \cellcolor[HTML]{C0C0C0}\color[HTML]{0000B0}3.33 & \multicolumn{1}{l|}{\color[HTML]{B00000}-8.86} & \cellcolor[HTML]{C0C0C0}\color[HTML]{B00000}-16.59 \\ \cline{2-16} \multicolumn{1}{|l|}{\multirow{-3}{*}{\rotatebox[origin=c]{90}{\tiny Billboard03}}} & LGS & \multicolumn{1}{l|}{\color[HTML]{00B000}1.60} & \multicolumn{1}{l|}{\cellcolor[HTML]{C0C0C0}\color[HTML]{00B000}1.01{\color[HTML]{000000} }} & \multicolumn{1}{l|}{\color[HTML]{00B000}1.92} & \cellcolor[HTML]{C0C0C0}\color[HTML]{B00000}-4.09 & \multicolumn{1}{l|}{\color[HTML]{0000B0}0.00} & \multicolumn{1}{l|}{\cellcolor[HTML]{C0C0C0}\color[HTML]{B00000}-2.27} & \multicolumn{1}{l|}{\color[HTML]{0000B0}0.00} & \cellcolor[HTML]{C0C0C0}\color[HTML]{B00000}-1.60 & \multicolumn{1}{l|}{\color[HTML]{00B000}-0.00} & \multicolumn{1}{l|}{\cellcolor[HTML]{C0C0C0}\color[HTML]{B00000}4.24} & \multicolumn{1}{l|}{\color[HTML]{0000B0}0.00} & \cellcolor[HTML]{C0C0C0}\color[HTML]{00B000}3.11 & \multicolumn{1}{l|}{\color[HTML]{B00000}-8.86} & \cellcolor[HTML]{C0C0C0}\color[HTML]{B00000}-27.77 \\ \hline\multicolumn{1}{|l|}{}                              & -   & \multicolumn{1}{l|}{\textbf{34.22}} & \multicolumn{1}{l|}{\cellcolor[HTML]{C0C0C0}\color[HTML]{000000}-6.53{\color[HTML]{000000} }} & \multicolumn{1}{l|}{\textbf{24.71}} & \cellcolor[HTML]{C0C0C0}\color[HTML]{000000}-6.40 & \multicolumn{1}{l|}{\textbf{25.01}} & \multicolumn{1}{l|}{\cellcolor[HTML]{C0C0C0}\color[HTML]{000000}1.80} & \multicolumn{1}{l|}{\textbf{27.83}} & \cellcolor[HTML]{C0C0C0}\color[HTML]{000000}0.99 & \multicolumn{1}{l|}{\textbf{18.62}} & \multicolumn{1}{l|}{\cellcolor[HTML]{C0C0C0}\color[HTML]{000000}6.11} & \multicolumn{1}{l|}{\textbf{15.73}} & \cellcolor[HTML]{C0C0C0}\color[HTML]{000000}5.32 & \multicolumn{1}{l|}{\textbf{47.75}} & \cellcolor[HTML]{C0C0C0}\color[HTML]{000000}0.86 \\ \cline{2-16} \multicolumn{1}{|l|}{}                              & ZM  & \multicolumn{1}{l|}{\color[HTML]{0000B0}0.00} & \multicolumn{1}{l|}{\cellcolor[HTML]{C0C0C0}\color[HTML]{00B000}-5.20{\color[HTML]{000000} }} & \multicolumn{1}{l|}{\color[HTML]{0000B0}0.00} & \cellcolor[HTML]{C0C0C0}\color[HTML]{00B000}-3.01 & \multicolumn{1}{l|}{\color[HTML]{0000B0}0.00} & \multicolumn{1}{l|}{\cellcolor[HTML]{C0C0C0}\color[HTML]{B00000}1.79} & \multicolumn{1}{l|}{\color[HTML]{0000B0}0.00} & \cellcolor[HTML]{C0C0C0}\color[HTML]{B00000}0.92 & \multicolumn{1}{l|}{\color[HTML]{00B000}-0.01} & \multicolumn{1}{l|}{\cellcolor[HTML]{C0C0C0}\color[HTML]{00B000}0.71} & \multicolumn{1}{l|}{\color[HTML]{B00000}0.00} & \cellcolor[HTML]{C0C0C0}\color[HTML]{B00000}5.32 & \multicolumn{1}{l|}{\color[HTML]{B00000}-7.97} & \cellcolor[HTML]{C0C0C0}\color[HTML]{B00000}0.03 \\ \cline{2-16} \multicolumn{1}{|l|}{\multirow{-3}{*}{\rotatebox[origin=c]{90}{\tiny Billboard04}}} & LGS & \multicolumn{1}{l|}{\color[HTML]{B00000}-0.14} & \multicolumn{1}{l|}{\cellcolor[HTML]{C0C0C0}\color[HTML]{B00000}-8.30{\color[HTML]{000000} }} & \multicolumn{1}{l|}{\color[HTML]{B00000}-0.71} & \cellcolor[HTML]{C0C0C0}\color[HTML]{B00000}-6.43 & \multicolumn{1}{l|}{\color[HTML]{0000B0}0.00} & \multicolumn{1}{l|}{\cellcolor[HTML]{C0C0C0}\color[HTML]{B00000}0.17} & \multicolumn{1}{l|}{\color[HTML]{0000B0}0.00} & \cellcolor[HTML]{C0C0C0}\color[HTML]{B00000}-2.32 & \multicolumn{1}{l|}{\color[HTML]{00B000}-0.01} & \multicolumn{1}{l|}{\cellcolor[HTML]{C0C0C0}\color[HTML]{00B000}5.80} & \multicolumn{1}{l|}{\color[HTML]{B00000}0.00} & \cellcolor[HTML]{C0C0C0}\color[HTML]{00B000}5.23 & \multicolumn{1}{l|}{\color[HTML]{B00000}-7.97} & \cellcolor[HTML]{C0C0C0}\color[HTML]{00B000}1.11 \\ \hline\multicolumn{1}{|l|}{}                              & -   & \multicolumn{1}{l|}{\textbf{31.01}} & \multicolumn{1}{l|}{\cellcolor[HTML]{C0C0C0}\color[HTML]{000000}-1.01{\color[HTML]{000000} }} & \multicolumn{1}{l|}{\textbf{25.10}} & \cellcolor[HTML]{C0C0C0}\color[HTML]{000000}-5.06 & \multicolumn{1}{l|}{\textbf{22.45}} & \multicolumn{1}{l|}{\cellcolor[HTML]{C0C0C0}\color[HTML]{000000}-0.21} & \multicolumn{1}{l|}{\textbf{22.17}} & \cellcolor[HTML]{C0C0C0}\color[HTML]{000000}-0.53 & \multicolumn{1}{l|}{\textbf{11.75}} & \multicolumn{1}{l|}{\cellcolor[HTML]{C0C0C0}\color[HTML]{000000}10.41} & \multicolumn{1}{l|}{\textbf{12.54}} & \cellcolor[HTML]{C0C0C0}\color[HTML]{000000}7.01 & \multicolumn{1}{l|}{\textbf{46.14}} & \cellcolor[HTML]{C0C0C0}\color[HTML]{000000}-4.10 \\ \cline{2-16} \multicolumn{1}{|l|}{}                              & ZM  & \multicolumn{1}{l|}{\color[HTML]{0000B0}0.00} & \multicolumn{1}{l|}{\cellcolor[HTML]{C0C0C0}\color[HTML]{B00000}-1.56{\color[HTML]{000000} }} & \multicolumn{1}{l|}{\color[HTML]{0000B0}0.00} & \cellcolor[HTML]{C0C0C0}\color[HTML]{00B000}-1.70 & \multicolumn{1}{l|}{\color[HTML]{0000B0}0.00} & \multicolumn{1}{l|}{\cellcolor[HTML]{C0C0C0}\color[HTML]{0000B0}-0.21} & \multicolumn{1}{l|}{\color[HTML]{0000B0}0.00} & \cellcolor[HTML]{C0C0C0}\color[HTML]{00B000}-0.53 & \multicolumn{1}{l|}{\color[HTML]{B00000}0.36} & \multicolumn{1}{l|}{\cellcolor[HTML]{C0C0C0}\color[HTML]{00B000}4.67} & \multicolumn{1}{l|}{\color[HTML]{B00000}0.03} & \cellcolor[HTML]{C0C0C0}\color[HTML]{0000B0}7.01 & \multicolumn{1}{l|}{\color[HTML]{B00000}-8.30} & \cellcolor[HTML]{C0C0C0}\color[HTML]{B00000}-9.69 \\ \cline{2-16} \multicolumn{1}{|l|}{\multirow{-3}{*}{\rotatebox[origin=c]{90}{\tiny Billboard05}}} & LGS & \multicolumn{1}{l|}{\color[HTML]{B00000}-0.20} & \multicolumn{1}{l|}{\cellcolor[HTML]{C0C0C0}\color[HTML]{00B000}-0.87{\color[HTML]{000000} }} & \multicolumn{1}{l|}{\color[HTML]{B00000}-0.24} & \cellcolor[HTML]{C0C0C0}\color[HTML]{00B000}-4.04 & \multicolumn{1}{l|}{\color[HTML]{0000B0}0.00} & \multicolumn{1}{l|}{\cellcolor[HTML]{C0C0C0}\color[HTML]{B00000}-1.01} & \multicolumn{1}{l|}{\color[HTML]{0000B0}0.00} & \cellcolor[HTML]{C0C0C0}\color[HTML]{B00000}-1.55 & \multicolumn{1}{l|}{\color[HTML]{B00000}0.36} & \multicolumn{1}{l|}{\cellcolor[HTML]{C0C0C0}\color[HTML]{00B000}9.74} & \multicolumn{1}{l|}{\color[HTML]{B00000}0.03} & \cellcolor[HTML]{C0C0C0}\color[HTML]{00B000}6.39 & \multicolumn{1}{l|}{\color[HTML]{B00000}-8.30} & \cellcolor[HTML]{C0C0C0}\color[HTML]{00B000}-2.17 \\ \hline\multicolumn{1}{|l|}{}                              & -   & \multicolumn{1}{l|}{\textbf{33.26}} & \multicolumn{1}{l|}{\cellcolor[HTML]{C0C0C0}\color[HTML]{000000}-2.31{\color[HTML]{000000} }} & \multicolumn{1}{l|}{\textbf{26.51}} & \cellcolor[HTML]{C0C0C0}\color[HTML]{000000}-9.50 & \multicolumn{1}{l|}{\textbf{24.04}} & \multicolumn{1}{l|}{\cellcolor[HTML]{C0C0C0}\color[HTML]{000000}-0.62} & \multicolumn{1}{l|}{\textbf{22.86}} & \cellcolor[HTML]{C0C0C0}\color[HTML]{000000}-0.23 & \multicolumn{1}{l|}{\textbf{11.47}} & \multicolumn{1}{l|}{\cellcolor[HTML]{C0C0C0}\color[HTML]{000000}9.71} & \multicolumn{1}{l|}{\textbf{8.11}} & \cellcolor[HTML]{C0C0C0}\color[HTML]{000000}8.08 & \multicolumn{1}{l|}{\textbf{64.95}} & \cellcolor[HTML]{C0C0C0}\color[HTML]{000000}-27.01 \\ \cline{2-16} \multicolumn{1}{|l|}{}                              & ZM  & \multicolumn{1}{l|}{\color[HTML]{0000B0}0.00} & \multicolumn{1}{l|}{\cellcolor[HTML]{C0C0C0}\color[HTML]{B00000}-4.12{\color[HTML]{000000} }} & \multicolumn{1}{l|}{\color[HTML]{0000B0}0.00} & \cellcolor[HTML]{C0C0C0}\color[HTML]{00B000}-5.87 & \multicolumn{1}{l|}{\color[HTML]{00B000}0.01} & \multicolumn{1}{l|}{\cellcolor[HTML]{C0C0C0}\color[HTML]{B00000}-1.22} & \multicolumn{1}{l|}{\color[HTML]{B00000}-0.09} & \cellcolor[HTML]{C0C0C0}\color[HTML]{00B000}-0.11 & \multicolumn{1}{l|}{\color[HTML]{0000B0}0.00} & \multicolumn{1}{l|}{\cellcolor[HTML]{C0C0C0}\color[HTML]{00B000}4.27} & \multicolumn{1}{l|}{\color[HTML]{B00000}0.04} & \cellcolor[HTML]{C0C0C0}\color[HTML]{00B000}5.40 & \multicolumn{1}{l|}{\color[HTML]{B00000}-0.08} & \cellcolor[HTML]{C0C0C0}\color[HTML]{00B000}-11.60 \\ \cline{2-16} \multicolumn{1}{|l|}{\multirow{-3}{*}{\rotatebox[origin=c]{90}{\tiny Billboard06}}} & LGS & \multicolumn{1}{l|}{\color[HTML]{B00000}-2.07} & \multicolumn{1}{l|}{\cellcolor[HTML]{C0C0C0}\color[HTML]{B00000}-4.06{\color[HTML]{000000} }} & \multicolumn{1}{l|}{\color[HTML]{B00000}-0.40} & \cellcolor[HTML]{C0C0C0}\color[HTML]{00B000}-6.65 & \multicolumn{1}{l|}{\color[HTML]{00B000}0.01} & \multicolumn{1}{l|}{\cellcolor[HTML]{C0C0C0}\color[HTML]{B00000}-1.98} & \multicolumn{1}{l|}{\color[HTML]{B00000}-0.09} & \cellcolor[HTML]{C0C0C0}\color[HTML]{B00000}-2.03 & \multicolumn{1}{l|}{\color[HTML]{0000B0}0.00} & \multicolumn{1}{l|}{\cellcolor[HTML]{C0C0C0}\color[HTML]{B00000}10.28} & \multicolumn{1}{l|}{\color[HTML]{B00000}0.04} & \cellcolor[HTML]{C0C0C0}\color[HTML]{00B000}7.73 & \multicolumn{1}{l|}{\color[HTML]{B00000}-0.08} & \cellcolor[HTML]{C0C0C0}\color[HTML]{00B000}-25.48 \\ \hline\multicolumn{1}{|l|}{}                              & -   & \multicolumn{1}{l|}{\textbf{30.73}} & \multicolumn{1}{l|}{\cellcolor[HTML]{C0C0C0}\color[HTML]{000000}-12.69{\color[HTML]{000000} }} & \multicolumn{1}{l|}{\textbf{20.56}} & \cellcolor[HTML]{C0C0C0}\color[HTML]{000000}-7.00 & \multicolumn{1}{l|}{\textbf{12.76}} & \multicolumn{1}{l|}{\cellcolor[HTML]{C0C0C0}\color[HTML]{000000}-1.01} & \multicolumn{1}{l|}{\textbf{9.43}} & \cellcolor[HTML]{C0C0C0}\color[HTML]{000000}-0.15 & \multicolumn{1}{l|}{\textbf{13.35}} & \multicolumn{1}{l|}{\cellcolor[HTML]{C0C0C0}\color[HTML]{000000}13.65} & \multicolumn{1}{l|}{\textbf{13.35}} & \cellcolor[HTML]{C0C0C0}\color[HTML]{000000}8.84 & \multicolumn{1}{l|}{\textbf{45.41}} & \cellcolor[HTML]{C0C0C0}\color[HTML]{000000}-33.17 \\ \cline{2-16} \multicolumn{1}{|l|}{}                              & ZM  & \multicolumn{1}{l|}{\color[HTML]{0000B0}0.00} & \multicolumn{1}{l|}{\cellcolor[HTML]{C0C0C0}\color[HTML]{0000B0}-12.69{\color[HTML]{000000} }} & \multicolumn{1}{l|}{\color[HTML]{B00000}-0.91} & \cellcolor[HTML]{C0C0C0}\color[HTML]{00B000}-4.84 & \multicolumn{1}{l|}{\color[HTML]{0000B0}0.00} & \multicolumn{1}{l|}{\cellcolor[HTML]{C0C0C0}\color[HTML]{00B000}-0.31} & \multicolumn{1}{l|}{\color[HTML]{0000B0}0.00} & \cellcolor[HTML]{C0C0C0}\color[HTML]{B00000}-0.17 & \multicolumn{1}{l|}{\color[HTML]{0000B0}0.00} & \multicolumn{1}{l|}{\cellcolor[HTML]{C0C0C0}\color[HTML]{00B000}6.12} & \multicolumn{1}{l|}{\color[HTML]{00B000}-0.08} & \cellcolor[HTML]{C0C0C0}\color[HTML]{00B000}8.68 & \multicolumn{1}{l|}{\color[HTML]{B00000}-0.33} & \cellcolor[HTML]{C0C0C0}\color[HTML]{00B000}-30.72 \\ \cline{2-16} \multicolumn{1}{|l|}{\multirow{-3}{*}{\rotatebox[origin=c]{90}{\tiny Billboard07}}} & LGS & \multicolumn{1}{l|}{\color[HTML]{00B000}0.87} & \multicolumn{1}{l|}{\cellcolor[HTML]{C0C0C0}\color[HTML]{00B000}-11.97{\color[HTML]{000000} }} & \multicolumn{1}{l|}{\color[HTML]{B00000}-0.43} & \cellcolor[HTML]{C0C0C0}\color[HTML]{00B000}-6.01 & \multicolumn{1}{l|}{\color[HTML]{0000B0}0.00} & \multicolumn{1}{l|}{\cellcolor[HTML]{C0C0C0}\color[HTML]{00B000}-0.95} & \multicolumn{1}{l|}{\color[HTML]{0000B0}0.00} & \cellcolor[HTML]{C0C0C0}\color[HTML]{B00000}-0.51 & \multicolumn{1}{l|}{\color[HTML]{0000B0}0.00} & \multicolumn{1}{l|}{\cellcolor[HTML]{C0C0C0}\color[HTML]{00B000}12.73} & \multicolumn{1}{l|}{\color[HTML]{00B000}-0.08} & \cellcolor[HTML]{C0C0C0}\color[HTML]{00B000}8.54 & \multicolumn{1}{l|}{\color[HTML]{B00000}-0.33} & \cellcolor[HTML]{C0C0C0}\color[HTML]{00B000}-32.83 \\ \hline\multicolumn{1}{|l|}{}                              & -   & \multicolumn{1}{l|}{\textbf{31.86}} & \multicolumn{1}{l|}{\cellcolor[HTML]{C0C0C0}\color[HTML]{000000}-3.03{\color[HTML]{000000} }} & \multicolumn{1}{l|}{\textbf{24.46}} & \cellcolor[HTML]{C0C0C0}\color[HTML]{000000}-2.93 & \multicolumn{1}{l|}{\textbf{37.18}} & \multicolumn{1}{l|}{\cellcolor[HTML]{C0C0C0}\color[HTML]{000000}0.78} & \multicolumn{1}{l|}{\textbf{38.59}} & \cellcolor[HTML]{C0C0C0}\color[HTML]{000000}-0.55 & \multicolumn{1}{l|}{\textbf{12.14}} & \multicolumn{1}{l|}{\cellcolor[HTML]{C0C0C0}\color[HTML]{000000}8.48} & \multicolumn{1}{l|}{\textbf{12.31}} & \cellcolor[HTML]{C0C0C0}\color[HTML]{000000}6.19 & \multicolumn{1}{l|}{\textbf{43.15}} & \cellcolor[HTML]{C0C0C0}\color[HTML]{000000}-8.01 \\ \cline{2-16} \multicolumn{1}{|l|}{}                              & ZM  & \multicolumn{1}{l|}{\color[HTML]{B00000}-0.11} & \multicolumn{1}{l|}{\cellcolor[HTML]{C0C0C0}\color[HTML]{00B000}-2.38{\color[HTML]{000000} }} & \multicolumn{1}{l|}{\color[HTML]{B00000}-1.02} & \cellcolor[HTML]{C0C0C0}\color[HTML]{B00000}-3.00 & \multicolumn{1}{l|}{\color[HTML]{B00000}-0.02} & \multicolumn{1}{l|}{\cellcolor[HTML]{C0C0C0}\color[HTML]{B00000}0.31} & \multicolumn{1}{l|}{\color[HTML]{B00000}-0.20} & \cellcolor[HTML]{C0C0C0}\color[HTML]{00B000}-0.29 & \multicolumn{1}{l|}{\color[HTML]{B00000}0.55} & \multicolumn{1}{l|}{\cellcolor[HTML]{C0C0C0}\color[HTML]{00B000}1.30} & \multicolumn{1}{l|}{\color[HTML]{B00000}0.29} & \cellcolor[HTML]{C0C0C0}\color[HTML]{0000B0}6.19 & \multicolumn{1}{l|}{\color[HTML]{B00000}-8.49} & \cellcolor[HTML]{C0C0C0}\color[HTML]{B00000}-8.93 \\ \cline{2-16} \multicolumn{1}{|l|}{\multirow{-3}{*}{\rotatebox[origin=c]{90}{\tiny Billboard08}}} & LGS & \multicolumn{1}{l|}{\color[HTML]{B00000}-3.28} & \multicolumn{1}{l|}{\cellcolor[HTML]{C0C0C0}\color[HTML]{00B000}-1.22{\color[HTML]{000000} }} & \multicolumn{1}{l|}{\color[HTML]{B00000}-0.00} & \cellcolor[HTML]{C0C0C0}\color[HTML]{00B000}-1.60 & \multicolumn{1}{l|}{\color[HTML]{B00000}-0.02} & \multicolumn{1}{l|}{\cellcolor[HTML]{C0C0C0}\color[HTML]{B00000}-0.26} & \multicolumn{1}{l|}{\color[HTML]{B00000}-0.20} & \cellcolor[HTML]{C0C0C0}\color[HTML]{B00000}-0.74 & \multicolumn{1}{l|}{\color[HTML]{B00000}0.55} & \multicolumn{1}{l|}{\cellcolor[HTML]{C0C0C0}\color[HTML]{B00000}9.21} & \multicolumn{1}{l|}{\color[HTML]{B00000}0.29} & \cellcolor[HTML]{C0C0C0}\color[HTML]{00B000}6.04 & \multicolumn{1}{l|}{\color[HTML]{B00000}-8.49} & \cellcolor[HTML]{C0C0C0}\color[HTML]{B00000}-8.58 \\ \hline\multicolumn{1}{|l|}{}                              & -   & \multicolumn{1}{l|}{\textbf{37.41}} & \multicolumn{1}{l|}{\cellcolor[HTML]{C0C0C0}\color[HTML]{000000}-1.11{\color[HTML]{000000} }} & \multicolumn{1}{l|}{\textbf{23.68}} & \cellcolor[HTML]{C0C0C0}\color[HTML]{000000}-6.75 & \multicolumn{1}{l|}{\textbf{14.29}} & \multicolumn{1}{l|}{\cellcolor[HTML]{C0C0C0}\color[HTML]{000000}-0.32} & \multicolumn{1}{l|}{\textbf{14.53}} & \cellcolor[HTML]{C0C0C0}\color[HTML]{000000}-3.39 & \multicolumn{1}{l|}{\textbf{19.81}} & \multicolumn{1}{l|}{\cellcolor[HTML]{C0C0C0}\color[HTML]{000000}1.74} & \multicolumn{1}{l|}{\textbf{15.84}} & \cellcolor[HTML]{C0C0C0}\color[HTML]{000000}2.61 & \multicolumn{1}{l|}{\textbf{61.73}} & \cellcolor[HTML]{C0C0C0}\color[HTML]{000000}-2.55 \\ \cline{2-16} \multicolumn{1}{|l|}{}                              & ZM  & \multicolumn{1}{l|}{\color[HTML]{0000B0}0.00} & \multicolumn{1}{l|}{\cellcolor[HTML]{C0C0C0}\color[HTML]{0000B0}-1.11{\color[HTML]{000000} }} & \multicolumn{1}{l|}{\color[HTML]{B00000}-4.76} & \cellcolor[HTML]{C0C0C0}\color[HTML]{00B000}-5.89 & \multicolumn{1}{l|}{\color[HTML]{0000B0}0.00} & \multicolumn{1}{l|}{\cellcolor[HTML]{C0C0C0}\color[HTML]{00B000}-0.22} & \multicolumn{1}{l|}{\color[HTML]{B00000}-0.00} & \cellcolor[HTML]{C0C0C0}\color[HTML]{00B000}-1.97 & \multicolumn{1}{l|}{\color[HTML]{00B000}-0.03} & \multicolumn{1}{l|}{\cellcolor[HTML]{C0C0C0}\color[HTML]{00B000}0.47} & \multicolumn{1}{l|}{\color[HTML]{00B000}-0.40} & \cellcolor[HTML]{C0C0C0}\color[HTML]{00B000}2.15 & \multicolumn{1}{l|}{\color[HTML]{B00000}-0.94} & \cellcolor[HTML]{C0C0C0}\color[HTML]{00B000}-1.50 \\ \cline{2-16} \multicolumn{1}{|l|}{\multirow{-3}{*}{\rotatebox[origin=c]{90}{\tiny Billboard09}}} & LGS & \multicolumn{1}{l|}{\color[HTML]{B00000}-0.18} & \multicolumn{1}{l|}{\cellcolor[HTML]{C0C0C0}\color[HTML]{B00000}-1.25{\color[HTML]{000000} }} & \multicolumn{1}{l|}{\color[HTML]{00B000}0.19} & \cellcolor[HTML]{C0C0C0}\color[HTML]{00B000}-5.83 & \multicolumn{1}{l|}{\color[HTML]{0000B0}0.00} & \multicolumn{1}{l|}{\cellcolor[HTML]{C0C0C0}\color[HTML]{B00000}-0.83} & \multicolumn{1}{l|}{\color[HTML]{B00000}-0.00} & \cellcolor[HTML]{C0C0C0}\color[HTML]{B00000}-3.72 & \multicolumn{1}{l|}{\color[HTML]{00B000}-0.03} & \multicolumn{1}{l|}{\cellcolor[HTML]{C0C0C0}\color[HTML]{B00000}2.17} & \multicolumn{1}{l|}{\color[HTML]{00B000}-0.40} & \cellcolor[HTML]{C0C0C0}\color[HTML]{00B000}2.34 & \multicolumn{1}{l|}{\color[HTML]{B00000}-0.94} & \cellcolor[HTML]{C0C0C0}\color[HTML]{B00000}-4.17 \\ \hline\multicolumn{1}{|l|}{}                              & -   & \multicolumn{1}{l|}{\textbf{27.89}} & \multicolumn{1}{l|}{\cellcolor[HTML]{C0C0C0}\color[HTML]{000000}0.09{\color[HTML]{000000} }} & \multicolumn{1}{l|}{\textbf{22.75}} & \cellcolor[HTML]{C0C0C0}\color[HTML]{000000}-4.51 & \multicolumn{1}{l|}{\textbf{10.22}} & \multicolumn{1}{l|}{\cellcolor[HTML]{C0C0C0}\color[HTML]{000000}0.29} & \multicolumn{1}{l|}{\textbf{12.68}} & \cellcolor[HTML]{C0C0C0}\color[HTML]{000000}0.49 & \multicolumn{1}{l|}{\textbf{14.08}} & \multicolumn{1}{l|}{\cellcolor[HTML]{C0C0C0}\color[HTML]{000000}1.21} & \multicolumn{1}{l|}{\textbf{11.97}} & \cellcolor[HTML]{C0C0C0}\color[HTML]{000000}0.15 & \multicolumn{1}{l|}{\textbf{42.88}} & \cellcolor[HTML]{C0C0C0}\color[HTML]{000000}-0.11 \\ \cline{2-16} \multicolumn{1}{|l|}{}                              & ZM  & \multicolumn{1}{l|}{\color[HTML]{B00000}-0.35} & \multicolumn{1}{l|}{\cellcolor[HTML]{C0C0C0}\color[HTML]{B00000}-0.20{\color[HTML]{000000} }} & \multicolumn{1}{l|}{\color[HTML]{00B000}0.46} & \cellcolor[HTML]{C0C0C0}\color[HTML]{00B000}-0.22 & \multicolumn{1}{l|}{\color[HTML]{B00000}-0.08} & \multicolumn{1}{l|}{\cellcolor[HTML]{C0C0C0}\color[HTML]{B00000}-0.22} & \multicolumn{1}{l|}{\color[HTML]{B00000}-0.00} & \cellcolor[HTML]{C0C0C0}\color[HTML]{B00000}0.47 & \multicolumn{1}{l|}{\color[HTML]{B00000}0.01} & \multicolumn{1}{l|}{\cellcolor[HTML]{C0C0C0}\color[HTML]{00B000}0.42} & \multicolumn{1}{l|}{\color[HTML]{B00000}0.08} & \cellcolor[HTML]{C0C0C0}\color[HTML]{00B000}0.11 & \multicolumn{1}{l|}{\color[HTML]{B00000}-0.40} & \cellcolor[HTML]{C0C0C0}\color[HTML]{B00000}-8.23 \\ \cline{2-16} \multicolumn{1}{|l|}{\multirow{-3}{*}{\rotatebox[origin=c]{90}{\tiny Truck}}} & LGS & \multicolumn{1}{l|}{\color[HTML]{B00000}-0.44} & \multicolumn{1}{l|}{\cellcolor[HTML]{C0C0C0}\color[HTML]{B00000}-0.39{\color[HTML]{000000} }} & \multicolumn{1}{l|}{\color[HTML]{B00000}-0.56} & \cellcolor[HTML]{C0C0C0}\color[HTML]{B00000}-4.62 & \multicolumn{1}{l|}{\color[HTML]{B00000}-0.08} & \multicolumn{1}{l|}{\cellcolor[HTML]{C0C0C0}\color[HTML]{B00000}0.05} & \multicolumn{1}{l|}{\color[HTML]{B00000}-0.00} & \cellcolor[HTML]{C0C0C0}\color[HTML]{B00000}-0.84 & \multicolumn{1}{l|}{\color[HTML]{B00000}0.01} & \multicolumn{1}{l|}{\cellcolor[HTML]{C0C0C0}\color[HTML]{00B000}0.77} & \multicolumn{1}{l|}{\color[HTML]{B00000}0.08} & \cellcolor[HTML]{C0C0C0}\color[HTML]{00B000}-0.18 & \multicolumn{1}{l|}{\color[HTML]{B00000}-0.40} & \cellcolor[HTML]{C0C0C0}\color[HTML]{B00000}-9.12 \\ \hline\end{tabular}}
\vskip -0.1 in
\caption{\small Performance of Z-Mask and LGS for each attack situation, for each task. White columns shows the results in the case without patch, whereas greyed columns shows the performance in the adversarial case. All the cells report the relative performance with respect to the undefended, non-adversarial case (in bold). The value is colored in red [green] if the defense is worsening [improving] the performance of the model, or blue if there is no effect.}\label{t:defenses}
\end{table}

\section{Conclusions} \label{s:conclusions}

This paper presented \NAME, a tool for the automatic generation of photo-realistic synthetic datasets that can be used to systematically evaluate the adversarial robustness of CNNs. The tool is based on the CARLA autonomous driving simulator, which overcomes the typical difficulties that arise while testing autonomous driving algorithms in various scenarios. At the current stage, \NAME~presents a few limitations. 

First of all, although the meshes used in Town10HD of CARLA are photo-realistic, there is a distributional shift between the CARLA-generated images and the datasets used to train the CNNs under test (COCO, CityScapes, Kitti). Furthermore, CARLA-generated ground-truth labels are often too fine-grained (especially for semantic segmentation).
This explains the poor performance of these models on the datasets generated with \NAME~with respect to the ones used for training.

Another issue with the simulated images is that the urban scenarios in CARLA towns are too small for a complete systematic evaluation of different scenarios, and the situations analysed in this paper do not cover all the possible patch attacks in the wild. Nonetheless, these problems will likely be fixed in the near future: simulators are becoming more and more photo-realistic, and developing a large enough city is just a matter of time. \NAME~ is general enough to be town-agnostic and can be easily extended to handle a large variety of different attack situations.
Another problem is that such a dataset generation process is computationally intensive and requires a significant amount of time, even with a powerful machines, especially when generating its own patches. This problem is alleviated by (i) providing a set of datasets with pre-computed patches for a set of CNNs spanning four different tasks, and (ii) providing custom patch and dataset generation upon email request to the authors. 

By changing the loss function used during the patch optimization process it is possible to change the attack objective, and, hence, its effect. This paper only presents white-box untargeted attacks. However, it is possible to extend the study to any kind of attack, e.g., targeted and black-box attacks.

The last open issue that must be mentioned concerns the development of a benchmark from the datasets generated by \NAME, which is left as a future work. The generation of datasets is only a first step towards the development of a benchmark for the evaluation of adversarial robustness and adversarial defense methods: more specific metrics need to be defined (probably focusing on an interesting subset of the classes), as well as a standardized evaluation platform, which will eventually lead to a leaderboard in adversarial detection, defense, and robustness.

\vfill
\pagebreak

\bibliographystyle{IEEEtran}
\bibliography{main}

\begin{thebibliography}{10}
\providecommand{\url}[1]{#1}
\csname url@samestyle\endcsname
\providecommand{\newblock}{\relax}
\providecommand{\bibinfo}[2]{#2}
\providecommand{\BIBentrySTDinterwordspacing}{\spaceskip=0pt\relax}
\providecommand{\BIBentryALTinterwordstretchfactor}{4}
\providecommand{\BIBentryALTinterwordspacing}{\spaceskip=\fontdimen2\font plus
\BIBentryALTinterwordstretchfactor\fontdimen3\font minus
  \fontdimen4\font\relax}
\providecommand{\BIBforeignlanguage}[2]{{%
\expandafter\ifx\csname l@#1\endcsname\relax
\typeout{** WARNING: IEEEtran.bst: No hyphenation pattern has been}%
\typeout{** loaded for the language `#1'. Using the pattern for}%
\typeout{** the default language instead.}%
\else
\language=\csname l@#1\endcsname
\fi
#2}}
\providecommand{\BIBdecl}{\relax}
\BIBdecl

\bibitem{2017arXiv171103938D}
A.~Dosovitskiy, G.~Ros, F.~Codevilla, A.~M. L{\'{o}}pez, and V.~Koltun,
  ``{CARLA:} an open urban driving simulator,'' in \emph{1st Annual Conference
  on Robot Learning, CoRL 2017, Mountain View, California, USA, November 13-15,
  2017, Proceedings}, ser. Proceedings of Machine Learning Research,
  vol.~78.\hskip 1em plus 0.5em minus 0.4em\relax {PMLR}, 2017, pp. 1--16.

\bibitem{9706854}
F.~Nesti, G.~Rossolini, S.~Nair, A.~Biondi, and G.~Buttazzo, ``Evaluating the
  robustness of semantic segmentation for autonomous driving against real-world
  adversarial patch attacks,'' in \emph{2022 IEEE/CVF Winter Conference on
  Applications of Computer Vision (WACV)}.\hskip 1em plus 0.5em minus
  0.4em\relax IEEE Computer Society, 2022, pp. 2826--2835.

\bibitem{biggio2013evasion}
B.~Biggio, I.~Corona, D.~Maiorca, B.~Nelson, N.~{\v{S}}rndi{\'c}, P.~Laskov,
  G.~Giacinto, and F.~Roli, ``Evasion attacks against machine learning at test
  time,'' in \emph{Joint European conference on machine learning and knowledge
  discovery in databases}.\hskip 1em plus 0.5em minus 0.4em\relax Springer,
  2013, pp. 387--402.

\bibitem{DBLP:journals/corr/SzegedyZSBEGF13}
C.~Szegedy, W.~Zaremba, I.~Sutskever, J.~Bruna, D.~Erhan, I.~J. Goodfellow, and
  R.~Fergus, ``Intriguing properties of neural networks,'' in \emph{2nd
  International Conference on Learning Representations, {ICLR} 2014, Banff, AB,
  Canada}, 2014.

\bibitem{kurakin_adversarial_2017}
A.~Kurakin, I.~J. Goodfellow, and S.~Bengio, ``Adversarial examples in the
  physical world,'' in \emph{5th International Conference on Learning
  Representations, {ICLR} 2017, Toulon, France, April 24-26, 2017, Workshop
  Track Proceedings}.\hskip 1em plus 0.5em minus 0.4em\relax OpenReview.net,
  2017.

\bibitem{brown_adversarial_2018}
T.~B. Brown, D.~Mané, A.~Roy, M.~Abadi, and J.~Gilmer, ``Adversarial
  {Patch},'' \emph{arXiv:1712.09665 [cs]}, May 2018.

\bibitem{pmlr-v80-athalye18b}
A.~Athalye, L.~Engstrom, A.~Ilyas, and K.~Kwok, ``Synthesizing robust
  adversarial examples,'' in \emph{Proceedings of the 35th International
  Conference on Machine Learning}, ser. Proceedings of Machine Learning
  Research, 2018, pp. 284--293.

\bibitem{moosavi2017universal}
S.-M. Moosavi-Dezfooli, A.~Fawzi, O.~Fawzi, and P.~Frossard, ``Universal
  adversarial perturbations,'' in \emph{Proceedings of the IEEE conference on
  computer vision and pattern recognition}, 2017, pp. 1765--1773.

\bibitem{dong2020benchmarking}
Y.~Dong, Q.-A. Fu, X.~Yang, T.~Pang, H.~Su, Z.~Xiao, and J.~Zhu, ``Benchmarking
  adversarial robustness on image classification,'' in \emph{Proceedings of the
  IEEE/CVF Conference on Computer Vision and Pattern Recognition}, 2020, pp.
  321--331.

\bibitem{croce2020robustbench}
F.~Croce, M.~Andriushchenko, V.~Sehwag, E.~Debenedetti, N.~Flammarion,
  M.~Chiang, P.~Mittal, and M.~Hein, ``Robustbench: a standardized adversarial
  robustness benchmark,'' in \emph{Thirty-fifth Conference on Neural
  Information Processing Systems Datasets and Benchmarks Track (Round 2)},
  2021.

\bibitem{mnistc}
N.~{Mu} and J.~{Gilmer}, ``{MNIST-C: A Robustness Benchmark for Computer
  Vision},'' \emph{arXiv e-prints}, p. arXiv:1906.02337, Jun. 2019.

\bibitem{biggio2022patches}
M.~{Pintor}, D.~{Angioni}, A.~{Sotgiu}, L.~{Demetrio}, A.~{Demontis},
  B.~{Biggio}, and F.~{Roli}, ``{ImageNet-Patch: A Dataset for Benchmarking
  Machine Learning Robustness against Adversarial Patches},'' \emph{arXiv
  e-prints}, p. arXiv:2203.04412, Mar. 2022.

\bibitem{jefferson_robust_2020}
B.~Jefferson and C.~O. Marrero, ``\BIBforeignlanguage{en}{Robust {Assessment}
  of {Real}-{World} {Adversarial} {Examples}},'' in
  \emph{\BIBforeignlanguage{en}{2020 {IEEE}/{CVF} {Conference} on {Computer}
  {Vision} and {Pattern} {Recognition} {Workshops} ({CVPRW})}}.\hskip 1em plus
  0.5em minus 0.4em\relax Seattle, WA, USA: IEEE, Jun. 2020, pp. 3442--3449.

\bibitem{2022arXiv220101850R}
G.~{Rossolini}, F.~{Nesti}, G.~{D'Amico}, S.~{Nair}, A.~{Biondi}, and
  G.~{Buttazzo}, ``{On the Real-World Adversarial Robustness of Real-Time
  Semantic Segmentation Models for Autonomous Driving},''
  \emph{arXiv:2201.01850}, 2022.

\bibitem{braunegg2020apricot}
A.~Braunegg, A.~Chakraborty, M.~Krumdick, N.~Lape, S.~Leary, K.~Manville,
  E.~Merkhofer, L.~Strickhart, and M.~Walmer, ``Apricot: A dataset of physical
  adversarial attacks on object detection,'' in \emph{European Conference on
  Computer Vision}.\hskip 1em plus 0.5em minus 0.4em\relax Springer, 2020, pp.
  35--50.

\bibitem{2022arXiv220307341R}
G.~{Rossolini}, F.~{Nesti}, F.~{Brau}, A.~{Biondi}, and G.~{Buttazzo},
  ``{Defending From Physically-Realizable Adversarial Attacks Through Internal
  Over-Activation Analysis},'' \emph{arXiv e-prints}, p. arXiv:2203.07341, Mar.
  2022.

\bibitem{naseer_local_2019}
M.~Naseer, S.~Khan, and F.~Porikli, ``Local gradients smoothing: Defense
  against localized adversarial attacks,'' in \emph{2019 {IEEE} Winter
  Conference on Applications of Computer Vision ({WACV})}, 2019.

\bibitem{hyperneuron}
K.~T. {Co}, L.~{Mu{\~n}oz-Gonz{\'a}lez}, L.~{Kanthan}, and E.~C. {Lupu},
  ``{Real-time Detection of Practical Universal Adversarial Perturbations},''
  \emph{arXiv e-prints}, p. arXiv:2105.07334, May 2021.

\bibitem{DBLP:conf/cvpr/CordtsORREBFRS16}
M.~Cordts, M.~Omran, S.~Ramos, T.~Rehfeld, M.~Enzweiler, R.~Benenson,
  U.~Franke, S.~Roth, and B.~Schiele, ``The cityscapes dataset for semantic
  urban scene understanding,'' in \emph{Conference on Computer Vision and
  Pattern Recognition {CVPR}}.\hskip 1em plus 0.5em minus 0.4em\relax {IEEE}
  Computer Society, 2016, pp. 3213--3223.

\bibitem{lin2014microsoft}
T.-Y. Lin, M.~Maire, S.~Belongie, J.~Hays, P.~Perona, D.~Ramanan,
  P.~Doll{\'a}r, and C.~L. Zitnick, ``Microsoft coco: Common objects in
  context,'' in \emph{European conference on computer vision}.\hskip 1em plus
  0.5em minus 0.4em\relax Springer, 2014, pp. 740--755.

\bibitem{KITTI}
A.~Geiger, P.~Lenz, and R.~Urtasun, ``Are we ready for autonomous driving? the
  kitti vision benchmark suite,'' in \emph{2012 IEEE Conference on Computer
  Vision and Pattern Recognition}.\hskip 1em plus 0.5em minus 0.4em\relax IEEE,
  2012, pp. 3354--3361.

\bibitem{ddrnet_paper}
Y.~Hong, H.~Pan, W.~Sun, and Y.~Jia, ``Deep dual-resolution networks for
  real-time and accurate semantic segmentation of road scenes,''
  \emph{arXiv:2101.06085}, 2021.

\bibitem{bisenet_paper}
C.~Yu, J.~Wang, C.~Peng, C.~Gao, G.~Yu, and N.~Sang, ``Bisenet: Bilateral
  segmentation network for real-time semantic segmentation,'' in
  \emph{Proceedings of the European Conference on Computer Vision
  (ECCV)}.\hskip 1em plus 0.5em minus 0.4em\relax Springer, 2018, pp. 325--341.

\bibitem{7485869}
S.~Ren, K.~He, R.~Girshick, and J.~Sun, ``Faster r-cnn: Towards real-time
  object detection with region proposal networks,'' \emph{IEEE Transactions on
  Pattern Analysis and Machine Intelligence}, vol.~39, no.~6, pp. 1137--1149,
  2017.

\bibitem{lin2017focal}
T.-Y. Lin, P.~Goyal, R.~Girshick, K.~He, and P.~Doll{\'a}r, ``Focal loss for
  dense object detection,'' in \emph{Proceedings of the IEEE international
  conference on computer vision (ICCV)}, 2017, pp. 2980--2988.

\bibitem{kim2022global}
D.~Kim, W.~Ga, P.~Ahn, D.~Joo, S.~Chun, and J.~Kim, ``Global-local path
  networks for monocular depth estimation with vertical cutdepth,'' \emph{arXiv
  preprint arXiv:2201.07436}, 2022.

\bibitem{Bhat_2021_CVPR}
S.~F. Bhat, I.~Alhashim, and P.~Wonka, ``Adabins: Depth estimation using
  adaptive bins,'' in \emph{Proceedings of the IEEE/CVF Conference on Computer
  Vision and Pattern Recognition (CVPR)}, June 2021, pp. 4009--4018.

\bibitem{li_stereo_2019}
P.~Li, X.~Chen, and S.~Shen, ``\BIBforeignlanguage{en}{Stereo {R}-{CNN} {Based}
  {3D} {Object} {Detection} for {Autonomous} {Driving}},'' in
  \emph{\BIBforeignlanguage{en}{2019 {IEEE}/{CVF} {Conference} on {Computer}
  {Vision} and {Pattern} {Recognition} ({CVPR})}}.\hskip 1em plus 0.5em minus
  0.4em\relax Long Beach, CA, USA: IEEE, Jun. 2019, pp. 7636--7644.

\end{thebibliography}

\end{document}